\documentclass[11pt]{article}

\usepackage[final]{acl}

\usepackage{times}
\usepackage{latexsym}

\usepackage[T1]{fontenc}

\usepackage[utf8]{inputenc}

\usepackage{microtype}

\usepackage{inconsolata}

\usepackage{graphicx}

\usepackage{multirow}
\usepackage{multicol}
\usepackage{makecell}
\usepackage{booktabs}
\usepackage{enumitem}
\usepackage{subfig}
\usepackage{subcaption}
\usepackage{wrapfig}

\title{Towards Fine-grained and Robust Detection of Human–AI Co-authorship}

\author{Ziruo Zhao \quad Jie Fu \quad Wendy Hui Wang \quad Zining Zhu \\
  Stevens Institute of Technology\\
  Hoboken, New Jersey, USA\\
  \texttt{\{zzhao83,jfu13,hwang4,zzhu41\}@stevens.edu}
}

\begin{document}
\maketitle
\begin{abstract}
The widespread adoption of large language models (LLMs) is reshaping how textual content is produced, complicating text provenance, and raising concerns for misinformation, plagiarism, and intellectual property. While prior work on AI-generated text detection has progressed, one-dimensional human-AI co-authorship scores lack signals directly relevant to ``AI involvement''. It remains unclear how adversarial paraphrasing distorts detection signals. 
In this work, we formulate a discrete-class text provenance detection framework. This framework allows an abstraction of realistic co-authorship scenarios into distinct degrees of AI involvement, differentiating the AI polishing on the surface level from the AI updating the discourse structures. This framework reveals the mechanisms of how adversarial paraphrasing blurs the decision boundaries for co-authorship detectors. To improve robustness of the detectors, we augment training data with adversarial paraphrases generated by the state-of-the-art paraphrasing attack model, exposing detectors to challenging meaning-preserving rewrites. Extensive experiments across multiple datasets and generation models demonstrate that adversarial paraphrasing substantially degrades detectors trained only on benign data, while adversarial data augmentation leads to consistent robustness gains and good cross-generator transferability. Overall, our findings highlight the value of discrete-class formulation and principled data augmentation for building reliable detectors of human-AI co-authorship.
\end{abstract}

\section{Introduction}

Large language models (LLMs) have become widely used as writing assistants, supporting a broad range of tasks such as drafting, editing, and paraphrasing text~\citep{liang2025quantifying,bhagat-hovy-2013-squibs}. As a result, many real-world documents are no longer purely human-written or fully machine-generated, but instead reflect varying degrees of human-AI co-authorship~\citep{paredes2025more,law2025web}. The shift introduces novel challenge, including the academic integrity, the tracing of information and the content provenance.  

Therefore, many recent works have sought to detect AI involvement in the writing. In practice, AI involvement in writing falls into diverse categories. Texts may undergo light AI-assisted editing, such as grammatical correction or stylistic refinement, or be substantially rewritten by language models while still originating from human-authored content. These intermediate cases blur the boundary between human-written and AI-generated text, recent work has started moving beyond binary detection toward fine-grained classification of AI involvement~\citep{abassy-etal-2024-llm,beyondbinary,su-etal-2025-haco,thai2026editlens}.

There are two unaddressed challenges. 

First, defining a continuous ``AI involvement'' score is inherently difficult, as the degree of AI involvement is ultimately a property of the authoring process, not directly recoverable from post-hoc detector outputs. While detectors like \citet{thai2026editlens} may leverage various signals from stylistic and geometric features to semantic and discourse-level cues, their output scores are not designed to map onto a one-dimensional ``AI involvement'' level. Content provenance requires a more direct signal showing the level of ``AI involvement''.

\begin{figure*}
    \centering
    \includegraphics[width=0.9\linewidth]{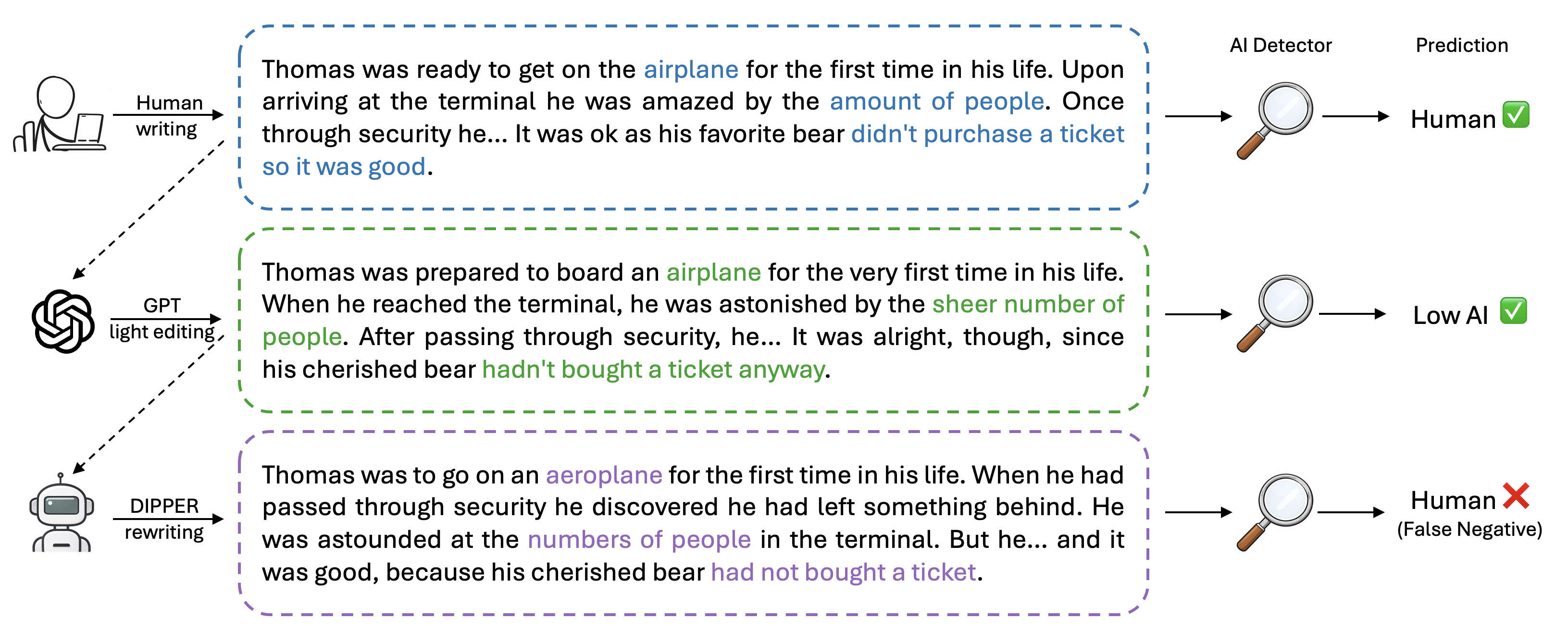}
    \caption{A ROCStories example showing that light GPT editing is correctly identified, while adversarial paraphrasing (DIPPER) preserves meaning but induces a false negative in an existing detector.}
    \label{fig:intro}
\end{figure*}

Second, adversarial paraphrasing poses a particularly realistic threat to AI-involved text detection~\citep{Sadasivan23,tripto-2024-ship}. Recent paraphrasing methods, such as DIPPER~\citep{Krishna23dipper}, are explicitly designed to generate meaning-preserving rewrites that substantially perturb lexical and syntactic patterns. As illustrated in Figure~\ref{fig:intro}, adversarial paraphrasing can cause texts with low AI involvement to be misclassified as human-written. This is a critical vulnerability of detectors, and the mechanisms behind this vulnerability remains largely unknown.

In this work, we tackle both challenges through a unified and controlled framework. 
Instead of trying to propose a continuous score that capture AI involvement, we take a step back and propose a four-class task formulation, distinguishing between multiple levels of AI involvements: ``pure-human'', ``low AI-involved'', ``high AI-involved'' and ``fully-AI''. 
This four-class setting corresponds to realistic scenarios of AI-assisted writing: after humans write ``pure-human'' class texts, one may use LLMs to polish the texts. This polish stays on the surface level, correcting typos and grammatical errors. This scenario is reflected by the ``low AI-involved'' class. The human user may use LLMs to develop the article, changing the argument structures and the discourse patterns. This scenario is reflected by the ``high AI-involved'' class. The human user may use LLMs to fill in the contents after briefly describing an idea or topic at the high level. The details in this scenario are generated by AI, and is reflected by the ``fully-AI'' class.
Therefore, this four-class setting provides signals that are directly relevant to the content provenance. The contents in ``low AI-involved'' texts should be attributed to the humans, while one could argue that the content details of the ``fully-AI'' texts can be attributed to the AI. For ``high AI-involved'' texts, one could argue that, for example, the contents attributed to humans while the discourse structures to the AIs. 

This formulation allows us to stress-test the detector's robustness when faced with adversarial paraphrasing. Under this setting, we study the impacts of adversarial paraphrasing on the co-authorship detectors. We validate that the adversarial paraphrasing obscures the syntactic and semantic signals used for detection, blurring the detection boundaries for traditional detection settings. To enhance the robustness, we augment the adversarially paraphrased data. We systematically evaluate robustness under increasing rewriting strength and cross-generator evaluation. 

Our contributions are summarized as follows:
\begin{itemize}[leftmargin=*,nosep]{
    \item We propose a four-class formulation for human–AI co-authorship detection, where we carefully control the different degrees of AI involvement. 
    \item We construct a benchmark dataset for fine-grained human–AI co-authorship detection under adversarial paraphrasing. We further validate that the dataset conforms to the intended AI-involved settings through quantitative analysis.
    \item We systematically analyze the impact of adversarial paraphrasing on both syntactic and semantic similarity, showing how meaning-preserving rewrites distort surface cues and exacerbate class overlap in fine-grained detection, thereby directly addressing the challenges posed by adversarial paraphrasing.
    \item We conduct comprehensive robustness evaluations across multiple datasets, generation models, and paraphrasing attacks, demonstrating consistent improvements in detection robustness under realistic rewriting scenarios.}
\end{itemize} 
Overall, our findings highlight the importance of fine-grained task formulation and data augmentation with adversarially paraphrased samples for building human-AI co-authorship detectors. Our framework and benchmark make a concrete step towards the data provenance of human-AI co-authored articles.

\section{Related Works}
\paragraph{AI-generated Text Detection Methods.}
Existing methods for AI-generated text detection can be broadly categorized into three paradigms.
(1) Zero-shot methods leverage model-internal statistics or generation artifacts such as likelihood-based signals or watermarking, but often require access to specific model internals, limiting applicability to third-party
text~\citep{detectgpt,detectllm,zhao2023protectinglanguagegenerationmodels,bao2024fastdetect}.
(2) Feature-based methods combine explicit statistical or linguistic features with lightweight classifiers, but remain sensitive to feature design and degrade under paraphrasing or stylistic variation~\citep{joulin17fasttext,Gehrmann19GLTR}.
(3) PLM-based methods fine-tune pretrained language models on labeled data for text classification, achieving strong in-distribution performance but exhibiting reduced robustness in out-of-distribution data~\citep{salazar-etal-2020-masked,rodriguez22cross}. To handle longer inputs, long-document transformers such as Longformer~\citep{longformer} extend standard PLMs with sparse attention mechanisms for efficient long-form text processing.

Overall, while existing detection methods perform well under benign conditions, they remain vulnerable to paraphrasing and distribution shifts, motivating the need for detection frameworks that explicitly consider adversarial rewriting and varying degrees of AI involvement.~\citep{pu23deepfake,guo2023closechatgpthumanexperts,li2023mage,li-etal-2024-spotting}

\paragraph{Human-AI Collaborative Text Detection.}
Recent work has begun to move beyond binary AI detection. Some studies treat human-AI collaboration as a distinct category~\citep{abassy-etal-2024-llm,beyondbinary,su-etal-2025-haco}, and another line of work considers humanized machine-generated text~\citep{zhang2024llmasacoauthormixedhumanwrittenmachinegenerated,masrour-etal-2025-damage}. A related direction focuses
on span or boundary detection, which identifies localized human-AI transitions within a document~\citep{kushnareva2024aigeneratedtextboundarydetection,li-etal-2024-spotting}. Most related to our work, \citet{thai2026editlens} propose EDITLENS, a regression model that quantifies the degree of AI editing using similarity-based metrics as a proxy for AI involvement. In contrast, our work operationalizes AI involvement directly through the generation process itself, using controlled rewriting scenarios as ground-truth supervision, and targets global involvement estimation under realistic full-document editing scenarios.

\paragraph{Paraphrasing attack.}
Recent studies have demonstrated that paraphrasing-based attacks pose a significant threat to AI-generated text detection. Paraphrasing-based attacks aim to preserve the original semantic content while altering surface-level lexical and syntactic patterns, thereby directly targeting the stylistic cues relied upon by many detectors. \citet{Sadasivan23,tripto-2024-ship} show that paraphrasing-based attacks can consistently evade a wide range of detection approaches.

Among existing methods, DIPPER~\citep{Krishna23dipper} is specifically designed to generate high-quality, meaning-preserving rewrites by jointly controlling semantic similarity and lexical diversity.  By selectively rewriting watermark-related tokens and adjusting sentence structure at the paragraph level, DIPPER generates natural and fluent text that more closely resembles human writing. While originally proposed to bypass watermarking-based detectors, our experiments show that DIPPER is also effective against other detection methods.

Several works further explore adversarial settings that explicitly model interactions between detectors and paraphrasers, including adversarial training or multi-round attack scenarios~\citep{hu23,Zhou24,universal,Shi23}. While effective, such frameworks typically require iterative detector--paraphraser interaction, resulting in substantial computational overhead and limiting scalability. Our work differs from their settings, focusing on meaning-preserving paraphrasing as a realistic and scalable threat model for studying detection robustness.


\section{Problem Setup}
This section defines the task formulation used throughout this work and characterizes how adversarial paraphrasing interacts with fine-grained human-AI co-authorship detection.

\subsection{4-Class Task Formulation}
\begin{table}[t]
  \centering
  \resizebox{\linewidth}{!}{
  \begin{tabular}{cccccc}
    \toprule
    \textbf{Generator} & \textbf{Data} & \textbf{Threshold} & \textbf{Low AI (\%)} & \textbf{High AI (\%)}\\
    \midrule
    \multirow{3}{*}{GPT} & XSum & 0.233 & 95.00 & 97.67 \\
     & arXiv & 0.247 & 94.61 & 97.06 \\
     & ROCStory & 0.605 & 94.99 & 97.83 \\
     \midrule
    \multirow{3}{*}{Claude} & XSum & 0.298 & 90.00 & 92.50 \\ 
     & arXiv & 0.366 & 94.61 & 91.18 \\
     & ROCStory & 0.678 & 94.99 & 98.83 \\
    \bottomrule
  \end{tabular}}
  \caption{Evaluation of surface-level separation on AI-paraphrased data using normalized Levenshtein distance. The percentages indicate the proportion of samples classified as low/high-AI involvement under each threshold.}
  \label{tab:levenshtein}
\end{table}

Instead of framing text detection as a binary classification problem, we formulate a four-class task to capture different degrees of AI involvement in writing: Human-written, Low AI-involved, High AI-involved, and Fully AI-generated. Throughout this work, AI involvement refers to the use of practically deployed large language models in real-world applications, such as ChatGPT and similar systems.

\textbf{Human-written} text is authored entirely by humans. \textbf{Low AI-involved} text corresponds to scenarios where AI assistance is limited to surface-level editing, such as refining lexical choice or adjusting local sentence form, while preserving the original discourse organization, and factual content. This setting aligns with common real-world applications such as grammar checking or light stylistic polishing tools.
\textbf{High AI-involved} text reflects a stronger form of assistance in which AI is entrusted with higher-level linguistic organization~\citep{namuduri2025qudsimquantifyingdiscoursesimilarities,huber-niklaus-2025-clear}. Given a human-authored draft or factual input, the model is allowed to organize discourse, expression, and narrative structure while preserving the underlying factual content. In this setting, AI contributes to structuring and re-articulating the text, rather than merely editing its surface form, which reflects common real-world scenarios in which large language models assist humans in drafting and reorganizing text.
For the \textbf{Fully AI-generated} class, a text is produced entirely by an AI model given only a writing direction or topic, with the model independently determining both the organization of content and its linguistic realization.

In this work, to simulate the defined AI-involved classes, labels are assigned based on the prompt-guided generation process rather than post-hoc similarity thresholds. Detailed descriptions of the data generation procedures are provided in Appendix~\ref{app:prompts}. To validate that the resulting categories exhibit the intended surface-level distinctions, we compute surface similarity measures between human-written text and its Low and High AI-involved counterparts. Table~\ref{tab:levenshtein} reports the proportion of samples categorized as Low or High AI involvement under different similarity thresholds across datasets, showing that the two classes remain consistently separable in terms of surface form.

In addition, Appendix~\ref{app:similarity} presents visualizations of both syntactic and semantic similarity metrics, further confirming that the generated data aligns with the intended distinctions between AI involvement levels. By explicitly modeling different degrees of AI involvement, our formulation moves beyond a binary authorship setting and enables a finer-grained analysis of detection behavior under varying levels of human–AI co-authorship. This distinction is particularly important in realistic writing scenarios, where human and AI contributions are often intertwined at different intensities.

\subsection{Impact of Adversarial Paraphrasing on Detection}
\begin{figure*}[t]
  \includegraphics[width=1\linewidth]{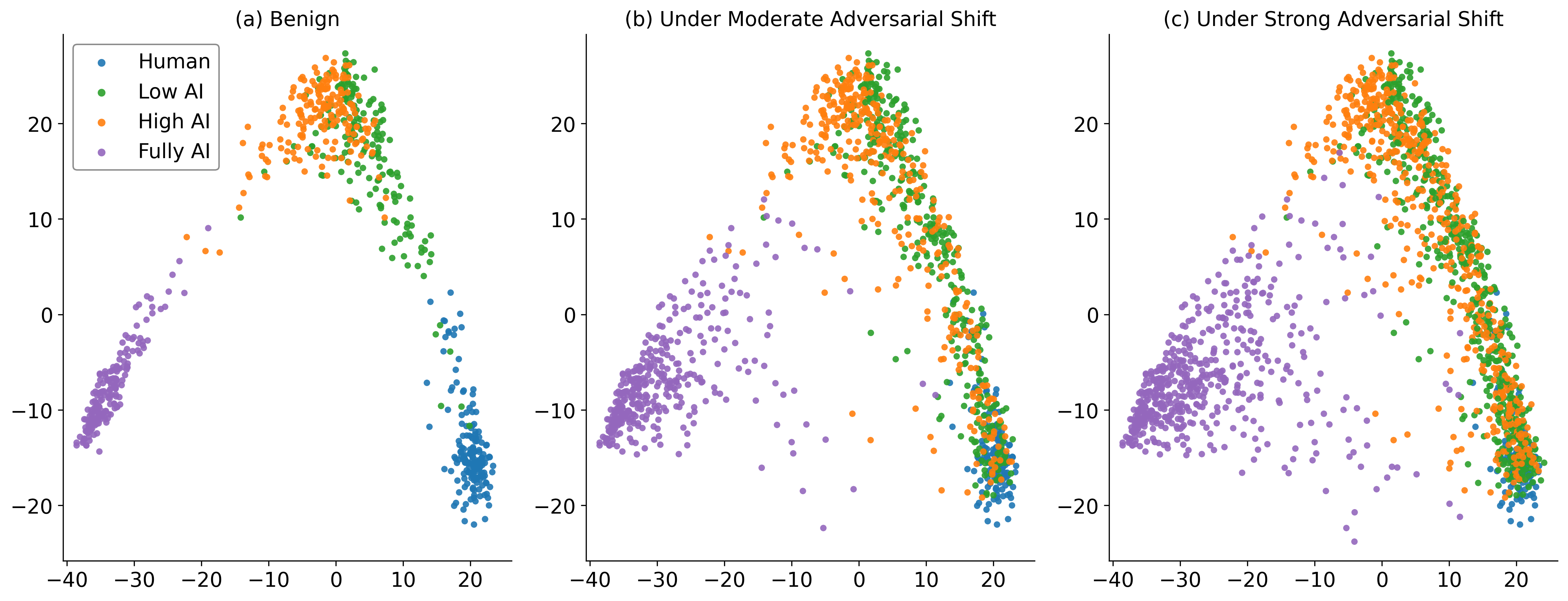}
  \caption {PCA visualization of document representations under increasing adversarial paraphrasing, with stronger paraphrases inducing shifts and overlap among AI-involved classes in representation space.}\label{pca_on_benign}
\end{figure*}
To analyze adversarial paraphrasing within the proposed four-class formulation, we first clarify its role with respect to class semantics. Adversarial paraphrasing operates as a meaning-preserving transformation applied to an existing document, designed to perturb surface-level cues exploited by detectors while preserving the underlying content and writing process. In our formulation, class labels are determined by the degree of AI involvement in the original generation process rather than by surface similarity or representation proximity. Consequently, although adversarial paraphrasing can substantially perturb lexical, syntactic, and stylistic patterns, it is treated within our formulation as preserving the original level of AI involvement, rather than introducing a new source of authorship.

To examine how adversarial paraphrasing affects detection behavior, we analyze document representations from a Longformer-based four-class detector trained solely on benign data, projected into two dimensions via PCA. Figure~\ref{pca_on_benign} presents three settings with increasing adversarial rewriting strength: (a) benign only, (b) combined with moderate adversarial samples, and (c) combined with stronger adversarial samples. 
Under the benign setting, representations form relatively compact and ordered clusters reflecting a continuous human–AI spectrum. As adversarial samples are introduced, this structure is progressively disrupted. In particular, Low and High AI-involved representations become widely dispersed along the "human-to-high AI'' continuum rather than shifting monotonically. This dispersion leads to substantial overlap between neighboring classes. 
In contrast, the Fully AI-generated class primarily exhibits cluster expansion, resulting in a broader but still coherent distribution. This asymmetric behavior indicates that adversarial paraphrasing induces non-directional variation within the human-authored regime rather than producing a new, more AI-involved category.

Overall, these observations show that adversarial paraphrasing causes systematic distributional shifts in representation space while preserving class semantics. This distinction explains why paraphrased samples retain their original labels yet remain challenging for detectors trained on benign data, motivating the need for training strategies that explicitly account for adversarial rewriting.

\section{Methodology}
Based on the problem formulation and challenges described above, we propose a supervised detection framework that incorporates adversarial paraphrasing during training to improve robustness under a fine-grained multi-class formulation.

\subsection{Augmenting Data with Adversarial Paraphrasing}
To enhance robustness against adversarial rewriting, we augment the training data with adversarially paraphrased samples. We adopt DIPPER~\citep{Krishna23dipper} as our primary paraphrasing method, as it produces high-quality, meaning-preserving rewrites that are designed to disrupt surface-level and structural cues exploited by pretrained language model–based detectors. Other paraphrasing approaches were explored but found to be less effective in generating reliable adversarial examples under our multi-class formulation.

Adversarial paraphrasing introduces substantial representational variation beyond the base four-class dataset, reducing over-reliance on generator-specific or surface-level features. While paraphrased documents may appear more human-like at the surface level, the rewriting process is a post-hoc transformation applied to existing content rather than a new instance of AI-assisted generation.

Consistent with our formulation, adversarially paraphrased samples retain the same class labels as their original inputs and are not treated as a separate category. This design preserves the semantics of AI involvement defined by the writing process, enabling systematic evaluation of detector robustness under increasing levels of adversarial rewriting.

\subsection{Training Detection Classifier}
Our study focuses on robustness and co-authorship detection under adversarial rewriting. Prior work~\citep{Krishna23dipper} has shown that paraphrasing-based attacks can effectively evade zero-shot and heuristic detectors; and such approaches are not designed to support controlled multi-class formulations. Therefore, we adopt a supervised PLM-based detection framework that enables explicit multi-class modeling and controlled robustness evaluation under adversarial rewriting.

We use Longformer~\citep{longformer} as the backbone encoder. Longformer replaces full self-attention with sparse attention patterns, allowing efficient modeling of long-range dependencies in long text. Given an input document, the encoder produces contextualized token representations in the final hidden layer.

We obtain a fixed-length document representation by applying mean pooling over token embeddings weighted by the attention mask. This representation is passed to a lightweight classification head consisting of a fully connected layer followed by a softmax function to predict the target label. Detection is formulated as a 4-class classification task and trained by minimizing cross-entropy loss.

\section{Benchmark Dataset Construction}
To enable systematic evaluation under the proposed four-class formulation, we construct a benchmark dataset using controlled, prompt-guided generation.\footnote{\url{https://github.com/zhaoziruo/CoAuthDetect}} This design supports fine-grained analysis by disentangling surface-level rewriting from underlying writing processes and enabling systematic examination of detector behavior.

\subsection{Human-written Data Sourcing}
We collect human-written texts from three data sources covering diverse writing domains:
\begin{enumerate}[leftmargin=*,nosep]
    \item News writing, including 600 BBC articles from the XSum dataset \citep{xsum};  
    \item Academic writing, consisting of 204 paper introductions collected from arXiv computer science categories published between 2020 and 2021;
    \item Story writing, including 599 five-sentence stories from the ROCStories corpus \citep{rocstory}.
\end{enumerate}

\subsection{Generation of AI-involved Classes}
For AI-involved classes generation, we use two large language models: GPT-4o \citep{gpt4o} and Claude Sonnet 4 \citep{claude4}.
For each human-written document, we generate machine-written variants using three prompting strategies that reflect increasing levels of AI involvement.
Low-AI prompts restrict the model to sentence-by-sentence rewriting with minimal surface-level edits.
High-AI prompts provide the full original text and allow more flexible rewriting, including sentence restructuring and stylistic variation.
Fully-AI prompts supply only a brief prefix of the original document and require the model to generate a complete continuation. Detailed prompts are provided in Appendix~\ref{app:prompts}.

\subsection{Adversarial Paraphrasing}
To further construct adversarial examples, we apply DIPPER~\citep{Krishna23dipper} to generate paraphrased versions of machine-written texts. Following the analysis in the original DIPPER study, we control paraphrasing strength using lexical diversity $L{=}60$ and order diversity $O{=}60$, which provides a balanced trade-off between surface-level variation and semantic preservation. DIPPER-based paraphrasing is applied single and two times to samples from three AI-involved classes.
In addition, we construct held-out adversarial test sets using GPT-5.1~\citep{gpt5.1}. This model is not used during training or data augmentation and serves solely for robustness evaluation.

\section{Experiments}

\begin{table*}[t]
  \centering
  \resizebox{0.8\linewidth}{!}{
  \begin{tabular}{clccc|lccc}
    \toprule
    \multirow{2}{*}{\textbf{Data}} 
    & \multicolumn{4}{c|}{\textbf{GPT-4o}} & \multicolumn{4}{c}{\textbf{Claude Sonnet 4}}\\
    \cmidrule(lr){2-9}
     & \textbf{Method} & \textbf{Acc} & \textbf{Macro-F1} & \textbf{AUC}
    & \textbf{Method} & \textbf{Acc} & \textbf{Macro-F1} & \textbf{AUC} \\
    \midrule

    \multirow{5}{*}{\textbf{XSum}}
      & Linguistic  & 0.8694 & 0.8682 & 0.9694 
      & Linguistic  & 0.9472 & 0.9471 & 0.9904 \\
      & Perplexity  & 0.5250 & 0.5054 & 0.7836 
      & Perplexity  & 0.5722 & 0.5561 & 0.8253 \\
      & BERT        & 0.9387 & 0.9388 & 0.9942 
      & BERT        & 0.9526 & 0.9525 & 0.9946 \\
      & RoBERTa     & 0.9716 & 0.9716 & 0.9995 
      & RoBERTa     & 0.9638 & 0.9638 & 0.9974 \\
      & Ours  & \textbf{0.9805} & \textbf{0.9805} & \textbf{0.9998} 
      & Ours  & \textbf{0.9749} & \textbf{0.9729} & \textbf{0.9979} \\
    \midrule
    \multirow{5}{*}{\textbf{ROCStory}}
      & Linguistic  & 0.8500 & 0.8489 & 0.9723 
      & Linguistic  & 0.8806 & 0.8808 & 0.9858 \\
      & Perplexity  & 0.5944 & 0.5914 & 0.8531 
      & Perplexity  & 0.6778 & 0.6733 & 0.8968 \\
      & BERT        & 0.9605 & 0.9605  & 0.9995
      & BERT & 0.9944 & 0.9945 & 0.9999 \\
      & RoBERTa     & 0.9577 & 0.9577 & 0.9992 
      & RoBERTa     & 0.9861 & 0.9861 & 0.9999 \\
      & Ours  & \textbf{0.9666} & \textbf{0.9665} & \textbf{0.9996}
      & Ours  & \textbf{0.9972} & \textbf{0.9972} & \textbf{1.0000} \\
    \midrule
    \multirow{5}{*}{\textbf{arXiv}}
      & Linguistic  & 0.5200 & 0.4438 & 0.8059 
      & Linguistic  & 0.7600 & 0.7464 & 0.9351 \\
      & Perplexity  & 0.5650 & 0.5657 & 0.7988 
      & Perplexity  & 0.6300 & 0.6253 & 0.8751 \\
      & BERT        & 0.6131 & 0.5862 & 0.8774 
      & BERT        & 0.7337 & 0.7182 & 0.9520 \\
      & RoBERTa     & 0.5980 & 0.5791 & 0.8461 
      & RoBERTa     & 0.8844 & 0.8800 & 0.9847 \\
      & Ours  & \textbf{0.7286} & \textbf{0.7252} & \textbf{0.9382} 
      & Ours  & \textbf{0.9296} & \textbf{0.9287} & \textbf{0.9913} \\                   
      \bottomrule
  \end{tabular}}
  \caption{Detection performance of baseline methods and ours.}
  \label{tab:baseline}
\end{table*}

\subsection{Experimental Setup}
We evaluate our approach under a controlled experimental setup designed to study human-AI co-authorship and robustness under adversarial rewriting. All experiments follow the proposed 4-class classification formulation.

\noindent\textbf{Training Settings.}
We consider three training settings with increasing levels of adversarial data augmentation. In the \textbf{Benign} setting, the detector is trained on the original 4-class dataset. \textbf{Level-1} augments the Benign setting with single-step DIPPER paraphrases. \textbf{Level-2} further extends Level-1 with a cumulative mixture of single-step and two-step DIPPER paraphrases. Adversarial paraphrasing is applied only to AI-involved samples, with paraphrased samples assigned the same labels as their original texts. All models are trained using class-weighted cross-entropy loss to address class imbalance.

\noindent\textbf{Testing Settings.}
For each training setting, we evaluate the detector on test sets constructed under the same three adversarial levels, enabling cross-level robustness evaluation.

\noindent\textbf{Metrics.}
We report Accuracy, Macro F1 score and ROC-AUC for all experiments, with per-class F1 scores for detailed analysis. Predictions are obtained using a default decision threshold of 0.5. Accuracy reflects performance at a fixed operating point, while ROC-AUC provides a threshold-independent measure of detection performance.

\subsection{Performance Evaluation}
We first establish baseline performance under benign conditions to contextualize the human-AI co-authorship detection task. We evaluate a set of representative supervised baselines, including feature-based classifiers and pretrained language model (PLM) classifiers, under the same 4-class formulation as our main approach.

For feature-based methods, we consider a Linguistic baseline implemented with FastText~\citep{joulin17fasttext} and a perplexity-based detector that uses LLaMA-2 to compute token-level perplexity, followed by a Random Forest classifier trained on aggregated perplexity statistics. We also evaluate supervised PLM-based classifiers, including BERT (uncased)~\citep{bert}, RoBERTa~\citep{roberta}, and our Longformer-based model~\citep{longformer}, all fine-tuned under identical data splits and evaluation protocols to ensure fair comparison.

Table~\ref{tab:baseline} reports performance of different methods under the benign in-distribution condition, with results over six settings formed by three datasets and two generators. Overall, PLM-based classifiers achieve higher performance than feature-based methods. Among PLM models, RoBERTa consistently improves over BERT, and our model achieves the strongest overall performance across metrics, showing its effectiveness in modeling co-authorship cue information across diverse domains.

\subsection{Adversarial Training Evaluation}
\begin{table*}
  \centering
  \resizebox{0.95\linewidth}{!}{
      \begin{tabular}{c|c|ccc|c|c|ccc}
        \toprule
        \multicolumn{2}{c|}{\textbf{Settings}} & 
        \multicolumn{3}{c|}{\textbf{Benign Data Generator: GPT-4o}} 
        & \multicolumn{2}{c|}{\textbf{Settings}} & \multicolumn{3}{c}{\textbf{Benign Data Generator: Claude}}\\
        \cmidrule(lr){3-5} \cmidrule(lr){8-10}
           \textbf{Train} & \textbf{Test} 
           & \textbf{Acc} & \textbf{F1} & \textbf{AUC}
           & \textbf{Train}
           & \textbf{Test} & \textbf{Acc} & \textbf{F1} & \textbf{AUC}\\
        \midrule
        \multicolumn{10}{c}{\textbf{(a) Same generator and same attack.}}\\
        \midrule
        Benign  & \multirow{3}{*}{Benign}  & 0.9805 & 0.9805 & \textbf{0.9998}
                & Benign & \multirow{3}{*}{Benign} & \textbf{0.9749} & \textbf{0.9729} & \textbf{0.9979} \\
        Level-1 & & \textbf{0.9861} & \textbf{0.9861} & 0.9990
                & Level-1 & & 0.9694 & 0.9693 & 0.9948 \\
        Level-2 & & 0.9638 & 0.9636 & 0.9973 
                & Level-2 & & 0.9610 & 0.9609 & 0.9960 \\
        \midrule
        Benign  & \multirow{3}{*}{Level-1}  & 0.7695 & 0.7560 & 0.9386
                & Benign & \multirow{3}{*}{Level-1} & 0.6789 & 0.6796 & 0.8999 \\
        Level-1 & & \textbf{0.9475} & \textbf{0.9514} & \textbf{0.9947} 
                & Level-1 & & \textbf{0.9491} & \textbf{0.9504} & \textbf{0.9927} \\
        Level-2 & & 0.9396 & 0.9413 & 0.9920 
                & Level-2 & & 0.9332 & 0.9356 & 0.9926 \\
        \midrule
        Benign  & \multirow{3}{*}{Level-2}  & 0.6652 & 0.6367 & 0.8988
                & Benign & \multirow{3}{*}{Level-2} & 0.5373 & 0.5426 & 0.8326 \\
        Level-1 & & \textbf{0.9321} & \textbf{0.9398} & \textbf{0.9921}
                & Level-1 & & \textbf{0.9321} & \textbf{0.9367} & 0.9890 \\
        Level-2 & & 0.9244 & 0.9292 & 0.9890 
                & Level-2 & & 0.9143 & 0.9197 & \textbf{0.9913} \\
        \midrule
        \multicolumn{10}{c}{\textbf{(b) Transfer across generators}}\\
        \midrule
        Benign  & \multirow{3}{*}{Claude}  & \textbf{0.9387} & \textbf{0.9386} & \textbf{0.9960}
                & Benign & \multirow{3}{*}{GPT-4o} & 0.8997 & 0.8991 & \textbf{0.9933} \\
        Level-1 & & 0.9062 & 0.9043 & 0.9898
                & Level-1 & & \textbf{0.9078} & \textbf{0.9084} & 0.9825 \\
        Level-2 & & 0.8877 & 0.8849 & 0.9833 
                & Level-2 & & 0.8954 & 0.9010 & 0.9816 \\        
        \midrule
        \multicolumn{10}{c}{\textbf{(c) Transfer across adversarial paraphrasing attacks}}\\
        \midrule
        Benign  & \multirow{3}{*}{GPT-5.1} & 0.6212 & 0.5377 & 0.8413 
                & Benign & \multirow{3}{*}{GPT-5.1} & 0.4903 & 0.3694 & 0.7882 \\
        Level-1 & & 0.8691 & \textbf{0.8682} & \textbf{0.9662} 
                & Level-1 & & \textbf{0.8134} & \textbf{0.8038} & \textbf{0.9433} \\
        Level-2 & & \textbf{0.8719} & 0.8680 & 0.9554 
                & Level-2 & & 0.7911 & 0.7756 & 0.9324 \\
        \bottomrule
      \end{tabular}}
  \caption{
  Results on the XSum dataset under different training and testing conditions. 
  Rows indicate increasing adversarial training levels based on DIPPER-based paraphrasing attack. 
  Columns compare detectors trained on benign data from different generators.
  GPT-5.1 serves as a held-out adversarial paraphrasing attack method.}
  \label{tab:xsum}
\end{table*}

Table~\ref{tab:xsum}(a) presents detection performance under different combinations of training and testing adversarial levels. We analyze how adversarial paraphrasing affects model robustness as the rewriting strength increases.

First, models trained only on benign data perform well under benign testing conditions but exhibit substantial performance degradation when evaluated on adversarial augmented test sets Level-1 and Level-2. This behavior indicates that detectors trained without adversarial exposure rely heavily on surface-level cues of the specific generator, which become unreliable once adversarial paraphrasing is introduced. 

In contrast, incorporating adversarial paraphrases during training yields consistent robustness gains. Models trained with Level-1 data maintain performance across all testing levels, suggesting exposure to single-step adversarial rewriting enables the detector to move beyond superficial lexical patterns and learn more stable signals that persist under paraphrasing.

To assess whether stronger augmentation is necessary, we further train models with Level-2 data including two-step DIPPER paraphrases. While Level-2 training preserves the robustness achieved by Level-1, improvements in accuracy and AUC are marginal, suggesting diminishing returns from increasing paraphrasing strength beyond a single step. Additional results on other datasets, reported in Appendix~\ref{additional_results} (Table~\ref{tab:rocstory} and \ref{tab:arxiv}) exhibit the same qualitative trends across domains.

\subsection{Transferability Study}

We further evaluate cross-generator generalization to assess whether detectors trained on one generation model can transfer to text generated by another. As summarized in Table~\ref{tab:xsum}(b), out-of-model performance is slightly lower than in-model evaluation, as expected given the distribution shift but remains competitive overall.

We then evaluate robustness under cross-adversarial attacks in Table~\ref{tab:xsum}(c). Although performance decreases on out-of-distribution adversarial samples generated by GPT-5.1, models trained with Level-1 augmentation consistently improve robustness in both training settings. AUC exceeds 0.94 in both cases, indicating strong cross-adversarial generalization.

\subsection{Ablation Study}
\begin{table}[t]
  \centering
  \resizebox{\linewidth}{!}{
  \begin{tabular}{lccc}
    \toprule
    \textbf{Augmentation (\%)} & \textbf{Accuracy} & \textbf{Macro-F1} & \textbf{AUC} \\
    \midrule
    0\%                  & 0.7695 & 0.7560 & 0.9386 \\
   25\%             & 0.9078 & 0.9127 & 0.9887 \\
   50\%             & 0.9348 & 0.9378 & 0.9893 \\
   75\%             & 0.9396 & 0.9427 & 0.9919 \\
   100\%          & \textbf{0.9475} & \textbf{0.9514} & \textbf{0.9947} \\
    \bottomrule
  \end{tabular}}
  \caption{
  Ablation study on the amounts of augmented data. All models are evaluated on the Level-1 test set.
  }
  \label{tab:ablation}
\end{table}

Next, we analyze the effect of mixing benign and adversarial samples during training. As shown in Table~\ref{tab:ablation}, increasing the proportion of adversarial samples leads to consistent performance improvements, confirming the benefit of adversarial exposure. Notably, the best results are obtained under the 100\% (Level-1) setting, where 100\% adversarial data is added on top of the benign data, combining benign samples with a diverse set of single-step adversarial paraphrases. This balanced training strategy prevents over-specialization to either purely benign or purely adversarial distributions, enabling the model to learn more robust and broadly applicable representations that generalize well across both benign and adversarial evaluation settings.

\section{Conclusions}

In this work, we study human–AI co-authorship detection under adversarial rewriting and propose a unified framework that explicitly models different degrees of AI involvement. By formulating detection as a four-class classification task, our framework moves beyond binary human-AI distinctions and enables fine-grained analysis of realistic AI-assisted writing scenarios. Extensive experiments show that adversarial data augmentation consistently improves robustness and generalization to out-of-distribution settings. We further find that a balanced mixture of benign and adversarial samples is sufficient to achieve robust performance. Overall, our results highlight the importance of principled task formulation, data augmentation design, and evaluation protocols for building reliable detectors in increasingly prevalent human–AI collaborative writing settings.


\section*{Limitations}
Our experiments involve a limited set of LLM generators and adversarial paraphrasers due to practical constraints such as API availability, cost, and output stability. Although we include cross-generator evaluation and held-out models to reduce overfitting to a single generator, fully covering the rapidly evolving landscape of large language models remains challenging.

In addition, real-world human–AI writing can involve more complex interaction patterns, such as multiple rounds of revision and interleaved human and AI contributions, which are difficult to simulate at scale. Our adversarial paraphrasing setup introduces diverse, meaning-preserving surface variations, but it does not fully capture interactive or human-driven editing dynamics. Exploring richer simulation frameworks for human-AI collaboration remains an important direction for future work.

\section*{Ethics Statement}
This work investigates the detection of human–AI co-authored text under varying degrees of AI involvement, including benign rewriting and adversarial paraphrasing. All experiments are conducted using publicly available datasets derived from human-written and model-generated text, and all data usage complies with the original dataset licenses. No new data involving human subjects is collected, and the datasets do not contain private or personally identifiable information.

This research has wide potential societal impact. First, our discrete AI involvement level classification framework provides support for text provenance detection, potentially informing discussions on copyright and authorship attribution. Second, we aim to improve the robustness and transparency of characterization of human-AI co-authored texts, a type of text that has become increasingly popular on the internet. At the same time, we acknowledge that detection and paraphrasing techniques could be misused by malicious users to, for example, evade detection systems, and the scope of this project does not include the defense to every potential attack. We emphasize that the proposed methods are intended for research purposes only and encourage any future research building upon this work to continue to consider the ethical implications.

\bibliography{custom}

@inproceedings{li2023mage,
  author       = {Yafu Li and
                  Qintong Li and
                  Leyang Cui and
                  Wei Bi and
                  Zhilin Wang and
                  Longyue Wang and
                  Linyi Yang and
                  Shuming Shi and
                  Yue Zhang},
  title        = {{MAGE:} Machine-generated Text Detection in the Wild},
  booktitle    = {Proceedings of the 62nd Annual Meeting of the Association for Computational
                  Linguistics (Volume 1: Long Papers), {ACL} 2024, Bangkok, Thailand,
                  August 11-16, 2024},
  pages        = {36--53},
  publisher    = {Association for Computational Linguistics},
  year         = {2024},
  url          = {https://doi.org/10.18653/v1/2024.acl-long.3}
}

@inproceedings{Krishna23dipper,
  author       = {Kalpesh Krishna and
                  Yixiao Song and
                  Marzena Karpinska and
                  John Wieting and
                  Mohit Iyyer},
  title        = {Paraphrasing evades detectors of AI-generated text, but retrieval
                  is an effective defense},
  booktitle    = {Advances in Neural Information Processing Systems 36: Annual Conference
                  on Neural Information Processing Systems 2023, NeurIPS 2023, New Orleans,
                  LA, USA, December 10 - 16, 2023},
  year         = {2023},
  url          = {http://papers.nips.cc/paper\_files/paper/2023/hash/575c450013d0e99e4b0ecf82bd1afaa4-Abstract-Conference.html}
}

@inproceedings{xsum,
    title = "Don{'}t Give Me the Details, Just the Summary! Topic-Aware Convolutional Neural Networks for Extreme Summarization",
    author = "Narayan, Shashi  and
      Cohen, Shay B.  and
      Lapata, Mirella",
    booktitle = "Proceedings of the 2018 Conference on Empirical Methods in Natural Language Processing",
    month = oct # "-" # nov,
    year = "2018",
    publisher = "Association for Computational Linguistics",
    url = "https://aclanthology.org/D18-1206/",
    pages = "1797--1807"
}

@inproceedings{rocstory,
    title = "A Corpus and Cloze Evaluation for Deeper Understanding of Commonsense Stories",
    author = "Mostafazadeh, Nasrin  and
      Chambers, Nathanael  and
      He, Xiaodong  and
      Parikh, Devi  and
      Batra, Dhruv  and
      Vanderwende, Lucy  and
      Kohli, Pushmeet  and
      Allen, James",
    booktitle = "Proceedings of the 2016 Conference of the North {A}merican Chapter of the Association for Computational Linguistics: Human Language Technologies",
    month = jun,
    year = "2016",
    publisher = "Association for Computational Linguistics",
    url = "https://aclanthology.org/N16-1098/",
    pages = "839--849"
}

@inproceedings{joulin17fasttext,
    title = "Bag of Tricks for Efficient Text Classification",
    author = "Joulin, Armand  and
      Grave, Edouard  and
      Bojanowski, Piotr  and
      Mikolov, Tomas",
    booktitle = "Proceedings of the 15th Conference of the {E}uropean Chapter of the Association for Computational Linguistics: Volume 2, Short Papers",
    month = apr,
    year = "2017",
    publisher = "Association for Computational Linguistics",
    url = "https://aclanthology.org/E17-2068/",
    pages = "427--431"
}

@inproceedings{Gehrmann19GLTR,
  author       = {Sebastian Gehrmann and
                  Hendrik Strobelt and
                  Alexander M. Rush},
  title        = {{GLTR:} Statistical Detection and Visualization of Generated Text},
  booktitle    = {Proceedings of the 57th Conference of the Association for Computational
                  Linguistics, {ACL} 2019, Florence, Italy, July 28 - August 2, 2019,
                  Volume 3: System Demonstrations},
  pages        = {111--116},
  publisher    = {Association for Computational Linguistics},
  year         = {2019},
  url          = {https://doi.org/10.18653/v1/p19-3019}
}

@article{Sadasivan23,
  author       = {Vinu Sankar Sadasivan and
                  Aounon Kumar and
                  Sriram Balasubramanian and
                  Wenxiao Wang and
                  Soheil Feizi},
  title        = {Can {AI}-Generated Text be Reliably Detected? Stress Testing {AI} Text Detectors Under Various Attacks},
  journal={Transactions on Machine Learning Research},
  issn={2835-8856},
  year={2025},
  url={https://openreview.net/forum?id=OOgsAZdFOt}
}

@inproceedings{hu23,
  title={{RADAR}: Robust {AI}-Text Detection via Adversarial Learning},
  author={Xiaomeng Hu and Pin-Yu Chen and Tsung-Yi Ho},
  booktitle={Thirty-seventh Conference on Neural Information Processing Systems},
  year={2023},
  url={https://openreview.net/forum?id=QGrkbaan79}
}

@inproceedings{Zhou24,
  author       = {Ying Zhou and
                  Ben He and
                  Le Sun},
  title        = {Humanizing Machine-Generated Content: Evading AI-Text Detection through
                  Adversarial Attack},
  booktitle    = {Proceedings of the 2024 Joint International Conference on Computational
                  Linguistics, Language Resources and Evaluation, {LREC/COLING} 2024,
                  20-25 May, 2024, Torino, Italy},
  pages        = {8427--8437},
  publisher    = {{ELRA} and {ICCL}},
  year         = {2024},
  url          = {https://aclanthology.org/2024.lrec-main.739}
}

@article{Shi23,
  title = "Red Teaming Language Model Detectors with Language Models",
    author = "Shi, Zhouxing  and
      Wang, Yihan  and
      Yin, Fan  and
      Chen, Xiangning  and
      Chang, Kai-Wei  and
      Hsieh, Cho-Jui",
    journal = "Transactions of the Association for Computational Linguistics",
    volume = "12",
    year = "2024",
    address = "Cambridge, MA",
    publisher = "MIT Press",
    url = "https://aclanthology.org/2024.tacl-1.10/",
    doi = "10.1162/tacl_a_00639",
    pages = "174--189"
}

@article{longformer,
  author       = {Iz Beltagy and
                  Matthew E. Peters and
                  Arman Cohan},
  title        = {Longformer: The Long-Document Transformer},
  journal      = {CoRR},
  volume       = {abs/2004.05150},
  year         = {2020},
  url          = {https://arxiv.org/abs/2004.05150}
}

@InProceedings{detectgpt,
  title = 	 {{D}etect{GPT}: Zero-Shot Machine-Generated Text Detection using Probability Curvature},
  author =       {Mitchell, Eric and Lee, Yoonho and Khazatsky, Alexander and Manning, Christopher D and Finn, Chelsea},
  booktitle = 	 {Proceedings of the 40th International Conference on Machine Learning},
  pages = 	 {24950--24962},
  year = 	 {2023},
  volume = 	 {202},
  series = 	 {Proceedings of Machine Learning Research},
  month = 	 {23--29 Jul},
  publisher =    {PMLR},
  url = 	 {https://proceedings.mlr.press/v202/mitchell23a.html}
}

@inproceedings{detectllm,
    title = "{D}etect{LLM}: Leveraging Log Rank Information for Zero-Shot Detection of Machine-Generated Text",
    author = "Su, Jinyan  and
      Zhuo, Terry  and
      Wang, Di  and
      Nakov, Preslav",
    booktitle = "Findings of the Association for Computational Linguistics: EMNLP 2023",
    pages = {12395--12412},
    month = "dec",
    year = "2023",
    publisher = "Association for Computational Linguistics",
    doi = {10.18653/v1/2023.findings-emnlp.827},
    url = "https://aclanthology.org/2023.findings-emnlp.827/"
}

@inproceedings{universal,
    title={Adversarial Paraphrasing: A Universal Attack for Humanizing {AI}-Generated Text},
    author={Yize CHENG and Vinu Sankar Sadasivan and Mehrdad Saberi and Shoumik Saha and Soheil Feizi},
    booktitle={The Thirty-ninth Annual Conference on Neural Information Processing Systems},
    year={2025},
    url={https://openreview.net/forum?id=fYjF9KIJd5}
}

@inproceedings{rodriguez22cross,
    title = "Cross-Domain Detection of {GPT}-2-Generated Technical Text",
    author = "Rodriguez, Juan Diego  and
      Hay, Todd  and
      Gros, David  and
      Shamsi, Zain  and
      Srinivasan, Ravi",
    booktitle = "Proceedings of the 2022 Conference of the North American Chapter of the Association for Computational Linguistics: Human Language Technologies",
    pages = {1213--1233},
    month = jul,
    year = "2022",
    publisher = "Association for Computational Linguistics",
    doi = {10.18653/v1/2022.naacl-main.88},
    url = "https://aclanthology.org/2022.naacl-main.88/"
}

@inproceedings{bert,
    title = "{BERT}: Pre-training of Deep Bidirectional Transformers for Language Understanding",
    author = "Devlin, Jacob  and
      Chang, Ming-Wei  and
      Lee, Kenton  and
      Toutanova, Kristina",
    booktitle = "Proceedings of the 2019 Conference of the North {A}merican Chapter of the Association for Computational Linguistics: Human Language Technologies, Volume 1 (Long and Short Papers)",
    month = jun,
    year = "2019",
    address = "Minneapolis, Minnesota",
    publisher = "Association for Computational Linguistics",
    url = "https://aclanthology.org/N19-1423/",
    doi = "10.18653/v1/N19-1423",
    pages = "4171--4186"
}

@article{roberta,
  author       = {Yinhan Liu and
                  Myle Ott and
                  Naman Goyal and
                  Jingfei Du and
                  Mandar Joshi and
                  Danqi Chen and
                  Omer Levy and
                  Mike Lewis and
                  Luke Zettlemoyer and
                  Veselin Stoyanov},
  title        = {{RoBERTa}: {A} Robustly Optimized {BERT} Pretraining Approach},
  journal      = {CoRR},
  volume       = {abs/1907.11692},
  year         = {2019},
  url          = {http://arxiv.org/abs/1907.11692}
}

@article{liang2025quantifying,
  title   = {Quantifying Large Language Model Usage in Scientific Papers},
  author  = {Liang, Weixin and Zhang, Yaohui and Wu, Zhengxuan and
             Lepp, Haley and Ji, Wenlong and Zhao, Xuandong and
             Cao, Hancheng and Liu, Sheng and He, Siyu and
             Huang, Zhi and Yang, Diyi and Potts, Christopher and
             Manning, Christopher D. and Zou, James},
  journal = {Nature Human Behaviour},
  volume  = {9},
  number  = {12},
  pages   = {2599--2609},
  year    = {2025},
  doi     = {10.1038/s41562-025-02273-8},
  url     = {https://doi.org/10.1038/s41562-025-02273-8}
}

@misc{paredes2025more,
  title={{More Articles Are Now Created by AI Than Humans}},
  author = {Paredes, Jose Luis and Smith, Ethan and Druck, Gregory and Benson, Bevin},
  url={https://graphite.io/five-percent/more-articles-are-now-created-by-ai-than-humans},
  year={2025}
}

@misc{law2025web,
    title = {{74\% of New Webpages Include AI Content (Study of 900k Pages)}},
    author = {Law, Ryan and Guan, Xibeijia and Soulo, Tim},
    year = {2025},
    url = {https://ahrefs.com/blog/what-percentage-of-new-content-is-ai-generated/}
}

@misc{gpt4o,
  title   = {{GPT-4o} System Card},
  author  = {{OpenAI}},
  year={2024},
  eprint={2410.21276},
  archivePrefix={arXiv},
  primaryClass={cs.CL},
  url={https://arxiv.org/abs/2410.21276} 
}

@misc{claude4,
    title = {Introducing Claude 4},
    author = {Anthropic},
    url={https://www.anthropic.com/news/claude-4},
    year = {2025}
}

@misc{gpt5.1,
    title={{GPT-5.1: A smarter, more conversational ChatGPT}},
    author = {OpenAI},
    url = {https://openai.com/index/gpt-5-1/},
    year = {2025}
}

@inproceedings{beyondbinary,
author = {Cheng, Zihao and Zhou, Li and Jiang, Feng and Wang, Benyou and Li, Haizhou},
title = {Beyond Binary: Towards Fine-Grained {LLM}-Generated Text Detection via Role Recognition and Involvement Measurement},
year = {2025},
isbn = {9798400712746},
publisher = {Association for Computing Machinery},
url = {https://doi.org/10.1145/3696410.3714770},
booktitle = {Proceedings of the ACM on Web Conference 2025},
pages = {2677--2688},
doi = {10.1145/3696410.3714770},
numpages = {12},
series = {WWW '25}
}

@inproceedings{tripto-2024-ship,
    title = "A Ship of Theseus: Curious Cases of Paraphrasing in {LLM}-Generated Texts",
    author = "Tripto, Nafis Irtiza  and
      Venkatraman, Saranya  and
      Macko, Dominik  and
      Moro, Robert  and
      Srba, Ivan  and
      Uchendu, Adaku  and
      Le, Thai  and
      Lee, Dongwon",
    booktitle = "Proceedings of the 62nd Annual Meeting of the Association for Computational Linguistics (Volume 1: Long Papers)",
    month = aug,
    year = "2024",
    publisher = "Association for Computational Linguistics",
    url = "https://aclanthology.org/2024.acl-long.357/",
    doi = "10.18653/v1/2024.acl-long.357",
    pages = "6608--6625"
}

@inproceedings{
bao2024fastdetect,
title={Fast-{D}etect{GPT}: Efficient Zero-Shot Detection of Machine-Generated Text via Conditional Probability Curvature},
author={Guangsheng Bao and Yanbin Zhao and Zhiyang Teng and Linyi Yang and Yue Zhang},
booktitle={The Twelfth International Conference on Learning Representations},
year={2024},
url={https://openreview.net/forum?id=Bpcgcr8E8Z}
}

@inproceedings{pu23deepfake,
  author       = {Jiameng Pu and
                  Zain Sarwar and
                  Sifat Muhammad Abdullah and
                  Abdullah Rehman and
                  Yoonjin Kim and
                  Parantapa Bhattacharya and
                  Mobin Javed and
                  Bimal Viswanath},
  title        = {Deepfake Text Detection: Limitations and Opportunities},
  booktitle    = {44th {IEEE} Symposium on Security and Privacy, {SP} 2023, San Francisco,
                  CA, USA, May 21-25, 2023},
  pages        = {1613--1630},
  publisher    = {{IEEE}},
  year         = {2023},
  url          = {https://doi.org/10.1109/SP46215.2023.10179387},
  doi          = {10.1109/SP46215.2023.10179387}
}

@inproceedings{salazar-etal-2020-masked,
    title = "Masked Language Model Scoring",
    author = "Salazar, Julian  and
      Liang, Davis  and
      Nguyen, Toan Q.  and
      Kirchhoff, Katrin",
    booktitle = "Proceedings of the 58th Annual Meeting of the Association for Computational Linguistics",
    month = jul,
    year = "2020",
    publisher = "Association for Computational Linguistics",
    url = "https://aclanthology.org/2020.acl-main.240/",
    doi = "10.18653/v1/2020.acl-main.240",
    pages = "2699--2712"
}

@misc{guo2023closechatgpthumanexperts,
      title={How Close is ChatGPT to Human Experts? Comparison Corpus, Evaluation, and Detection}, 
      author={Biyang Guo and Xin Zhang and Ziyuan Wang and Minqi Jiang and Jinran Nie and Yuxuan Ding and Jianwei Yue and Yupeng Wu},
      year={2023},
      eprint={2301.07597},
      archivePrefix={arXiv},
      primaryClass={cs.CL},
      url={https://arxiv.org/abs/2301.07597}, 
}

@article{bhagat-hovy-2013-squibs,
    title = "{S}quibs: What Is a Paraphrase?",
    author = "Bhagat, Rahul  and
      Hovy, Eduard",
    journal = "Computational Linguistics",
    volume = "39",
    number = "3",
    month = sep,
    year = "2013",
    address = "Cambridge, MA",
    publisher = "MIT Press",
    url = "https://aclanthology.org/J13-3001/",
    doi = "10.1162/COLI_a_00166",
    pages = "463--472"
}

@misc{zhao2023protectinglanguagegenerationmodels,
      title={Protecting Language Generation Models via Invisible Watermarking}, 
      author={Xuandong Zhao and Yu-Xiang Wang and Lei Li},
      year={2023},
      eprint={2302.03162},
      archivePrefix={arXiv},
      primaryClass={cs.CR},
      url={https://arxiv.org/abs/2302.03162}, 
}

@inproceedings{su-etal-2025-haco,
    title = "{HAC}o-{D}et: A Study Towards Fine-Grained Machine-Generated Text Detection under Human-{AI} Coauthoring",
    author = "Su, Zhixiong  and
      Wang, Yichen  and
      Wan, Herun  and
      Zhang, Zhaohan  and
      Luo, Minnan",
    booktitle = "Proceedings of the 63rd Annual Meeting of the Association for Computational Linguistics (Volume 1: Long Papers)",
    month = jul,
    year = "2025",
    publisher = "Association for Computational Linguistics",
    url = "https://aclanthology.org/2025.acl-long.1069/",
    doi = "10.18653/v1/2025.acl-long.1069",
    pages = "22015--22036"
}

@inproceedings{abassy-etal-2024-llm,
    title = "{LLM}-{D}etect{AI}ve: a Tool for Fine-Grained Machine-Generated Text Detection",
    author = "Abassy, Mervat  and
      Elozeiri, Kareem  and
      Aziz, Alexander  and
      Ta, Minh Ngoc  and
      Tomar, Raj Vardhan  and
      Adhikari, Bimarsha  and
      Ahmed, Saad El Dine  and
      Wang, Yuxia  and
      Mohammed Afzal, Osama  and
      Xie, Zhuohan  and
      Mansurov, Jonibek  and
      Artemova, Ekaterina  and
      Mikhailov, Vladislav  and
      Xing, Rui  and
      Geng, Jiahui  and
      Iqbal, Hasan  and
      Mujahid, Zain Muhammad  and
      Mahmoud, Tarek  and
      Tsvigun, Akim  and
      Aji, Alham Fikri  and
      Shelmanov, Artem  and
      Habash, Nizar  and
      Gurevych, Iryna  and
      Nakov, Preslav",
    booktitle = "Proceedings of the 2024 Conference on Empirical Methods in Natural Language Processing: System Demonstrations",
    month = nov,
    year = "2024",
    publisher = "Association for Computational Linguistics",
    url = "https://aclanthology.org/2024.emnlp-demo.35/",
    doi = "10.18653/v1/2024.emnlp-demo.35",
    pages = "336--343"
}

@inproceedings{li-etal-2024-spotting,
    title = "Spotting {AI}{'}s Touch: Identifying {LLM}-Paraphrased Spans in Text",
    author = "Li, Yafu  and
      Wang, Zhilin  and
      Cui, Leyang  and
      Bi, Wei  and
      Shi, Shuming  and
      Zhang, Yue",
    booktitle = "Findings of the Association for Computational Linguistics: ACL 2024",
    month = aug,
    year = "2024",
    publisher = "Association for Computational Linguistics",
    url = "https://aclanthology.org/2024.findings-acl.423/",
    doi = "10.18653/v1/2024.findings-acl.423",
    pages = "7088--7107"
}

@misc{kushnareva2024aigeneratedtextboundarydetection,
      title={{AI}-generated text boundary detection with {RoFT}}, 
      author={Laida Kushnareva and Tatiana Gaintseva and German Magai and Serguei Barannikov and Dmitry Abulkhanov and Kristian Kuznetsov and Eduard Tulchinskii and Irina Piontkovskaya and Sergey Nikolenko},
      year={2024},
      eprint={2311.08349},
      archivePrefix={arXiv},
      primaryClass={cs.CL},
      url={https://arxiv.org/abs/2311.08349}, 
}

@misc{zhang2024llmasacoauthormixedhumanwrittenmachinegenerated,
      title={{LLM}-as-a-{C}oauthor: Can Mixed Human-Written and Machine-Generated Text Be Detected?}, 
      author={Qihui Zhang and Chujie Gao and Dongping Chen and Yue Huang and Yixin Huang and Zhenyang Sun and Shilin Zhang and Weiye Li and Zhengyan Fu and Yao Wan and Lichao Sun},
      year={2024},
      eprint={2401.05952},
      archivePrefix={arXiv},
      primaryClass={cs.CL},
      url={https://arxiv.org/abs/2401.05952}, 
}

@inproceedings{
namuduri2025qudsimquantifyingdiscoursesimilarities,
title={{QUD}sim: Quantifying Discourse Similarities in {LLM}-Generated Text},
author={Ramya Namuduri and Yating Wu and Anshun Asher Zheng and Manya Wadhwa and Greg Durrett and Junyi Jessy Li},
booktitle={Second Conference on Language Modeling},
year={2025},
url={https://openreview.net/forum?id=zFz1BJu211}
}

@inproceedings{huber-niklaus-2025-clear,
    title = "{CLEAR}: A Comprehensive Linguistic Evaluation of Argument Rewriting by Large Language Models",
    author = "Huber, Thomas  and
      Niklaus, Christina",
    booktitle = "Findings of the Association for Computational Linguistics: EMNLP 2025",
    month = nov,
    year = "2025",
    publisher = "Association for Computational Linguistics",
    url = "https://aclanthology.org/2025.findings-emnlp.1065/",
    doi = "10.18653/v1/2025.findings-emnlp.1065",
    pages = "19548--19568",
    ISBN = "979-8-89176-335-7"
}

@inproceedings{masrour-etal-2025-damage,
    title = "{DAMAGE}: Detecting Adversarially Modified {AI} Generated Text",
    author = "Masrour, Elyas  and
      Emi, Bradley N.  and
      Spero, Max",
    booktitle = "Proceedings of the 1stWorkshop on GenAI Content Detection (GenAIDetect)",
    month = jan,
    year = "2025",
    publisher = "International Conference on Computational Linguistics",
    url = "https://aclanthology.org/2025.genaidetect-1.9/",
    pages = "120--133"
}

@inproceedings{
thai2026editlens,
title={Edit{L}ens: Quantifying the Extent of {AI} Editing in Text},
author={Katherine Thai and Bradley Emi and Elyas Masrour and Mohit Iyyer},
booktitle={The Fourteenth International Conference on Learning Representations},
year={2026},
url={https://openreview.net/forum?id=gOkitaPCfZ}
}

\appendix
\section{Prompt Examples}
\label{app:prompts}

We provide representative prompt examples used to construct light and heavy paraphrases, as well as fully generated text. 
For clarity, we present abstracted prompt templates and illustrate them using news articles (XSum), which are the primary domain analyzed in our experiments.
Across all datasets, prompts share the same rewriting objectives and constraints, with only minor adaptations for domain-specific writing style.
Adversarial paraphrases generated with GPT-5.1 are used solely for evaluation and are excluded from training.

\subsection{Low AI-involved}
This setting restricts the model to surface realization only, enforcing a fixed sentence plan and a strict one-to-one sentence mapping while preserving all factual content.

\begin{quote}
\textsc{Prompt.}\\
You are a professional news journalist. Rewrite the article sentence by sentence with a strict one-to-one mapping. 
Preserve all facts, entities, dates, and numerical values. 
Use a formal news style and keep sentence length within a small deviation from the original. 
Do not merge or split sentences. 
Return only the rewritten article.
\end{quote}

\subsection{High AI-involved}
This setting allows the model to perform generative sentence planning and surface realization under a fixed content plan, enabling substantial restructuring of sentences and paragraphs while preserving all factual content.

\begin{quote}
\textsc{Prompt.}\\
You are a creative news journalist. Heavily rewrite the article by restructuring sentences and paragraphs as needed. 
You may merge or split sentences and vary phrasing and discourse structure, while preserving all factual content. 
Keep the overall length within a moderate range of the original. 
Return only the rewritten article.
\end{quote}

\subsection{Fully AI-generated}
In this setting, the model autonomously performs content planning, determining how to expand a brief introduction into a complete article by selecting, organizing, and elaborating content.

\begin{quote}
\textsc{Prompt.}\\
You are a professional news journalist. Given a brief news introduction or summary, expand it into a complete news article. 
Add relevant background and context while maintaining a formal news style. 
Do not repeat the original introduction verbatim. 
Return only the generated article.
\end{quote}

\subsection{Adversarial Paraphrasing with GPT-5.1}

Adversarial paraphrasing evaluates robustness under a held-out generator by introducing surface variation while preserving meaning.

\begin{quote}
\textsc{Prompt.}\\
You are a professional editor. Paraphrase the following text to sound natural and fluent, as if written by a human, while preserving the original meaning and all factual details. 
Return only the paraphrased text.
\end{quote}


\section{Representation Analysis}
\label{app:tenclass}
\begin{figure*}[t]
    \centering
    \includegraphics[width=1\linewidth]{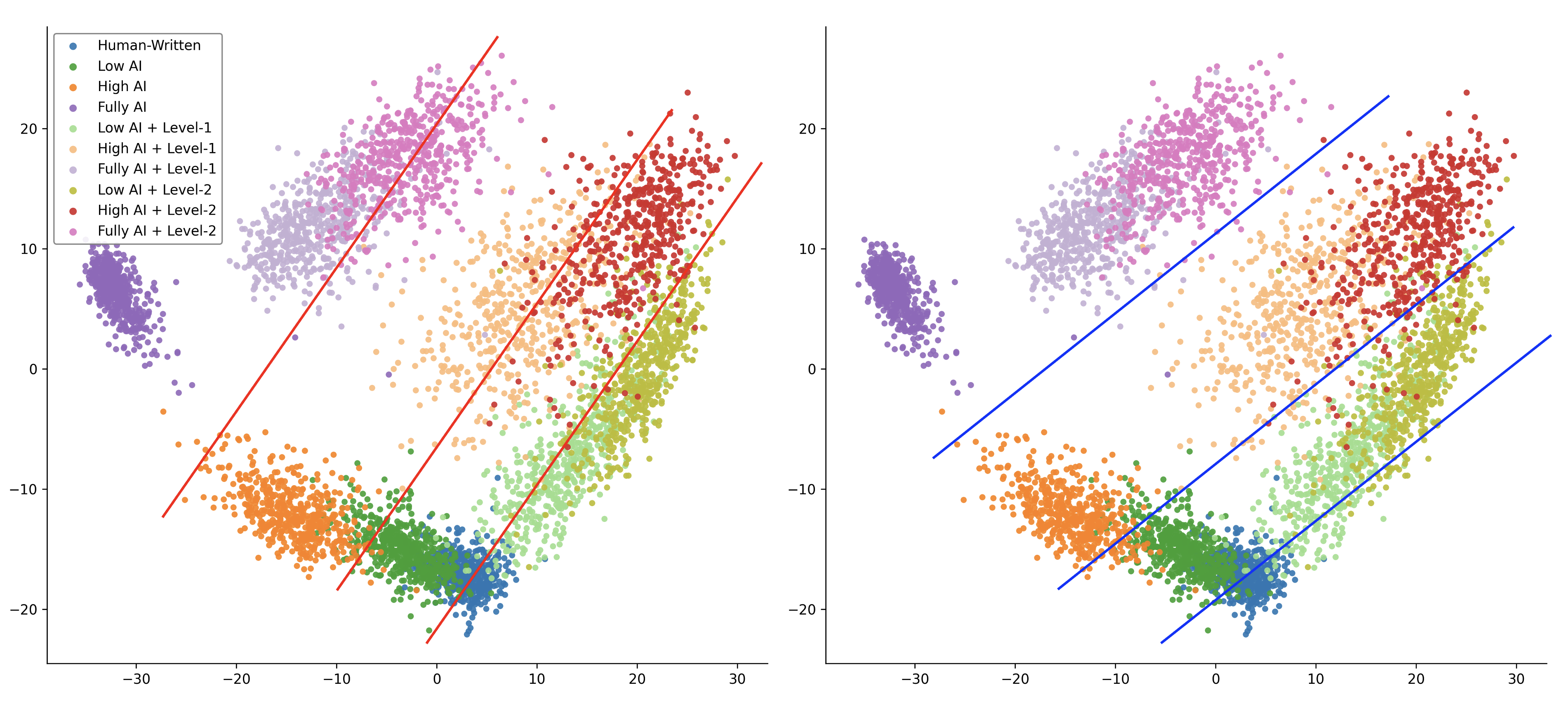}
    \caption{PCA visualization of document representations from a Longformer detector fine-tuned under a 10-class formulation. Each class corresponds to a benign or adversarial data variant treated as a separate label. The separation patterns illustrate that adversarial paraphrases occupy distinct directions in representation space rather than extending the benign AI-involvement continuum.}\label{fig:tenclass}
\end{figure*}

Figure~\ref{fig:tenclass} illustrates a PCA visualization of representations extracted from the fine-tuned Longformer model under a 10-class formulation on GPT-generated XSum data. Each class corresponds to a specific combination of AI-involvement level and adversarial strength, including benign samples, Level-1 and Level-2 DIPPER paraphrases, resulting in ten distinct data variants.

As shown in Fig.~\ref{fig:tenclass}(a), the learned representation space exhibits a clear directional structure. Samples generated directly by GPT and those produced via DIPPER paraphrasing tend to separate along different principal directions, indicating that adversarial paraphrasing induces systematic, directionally consistent shifts rather than random perturbations. This suggests that although DIPPER is designed as a paraphrasing attack model, it introduces its own stylistic regularities and does not fully emulate the diversity of genuine human-written text. Consequently, DIPPER-based samples form coherent clusters instead of smoothly interpolating within the human-writing distribution.

We further compare two alternative strategies for organizing class boundaries in the representation space. The red boundaries in Fig.~\ref{fig:tenclass}(a) attempt to reassign adversarially paraphrased samples across AI-involvement levels based on assumed paraphrasing strength (e.g., treating twice-paraphrased high-AI samples as low-AI). However, this strategy does not align well with the intrinsic geometry learned by the model: clusters intersect multiple red regions and fail to exhibit coherent separation. This mismatch indicates that paraphrasing strength alone is insufficient to redefine AI-involvement categories in the learned feature space.

In contrast, Fig.~\ref{fig:tenclass}(b) shows the blue-boundary formulation, which groups samples according to their original AI-involvement labels while allowing adversarial paraphrases to expand the surface realizations within each group. Under this formulation, clusters are more consistently aligned with decision boundaries, suggesting that mixing benign and adversarial samples encourages the model to learn broader, generator-agnostic representations rather than overfitting to paraphrase-specific artifacts. This observation provides qualitative support for our data augmentation strategy and helps explain why single-step adversarial training improves robustness without requiring aggressive multi-step paraphrasing.


\section{Quantitative Characterization}
\label{app:similarity}

We conduct quantitative analyses to validate the construction of benign AI-involvement degrees and to characterize how adversarial paraphrasing departs from this benign continuum. The analysis aims to distinguish natural variation induced by increasing AI involvement from additional surface-level perturbations introduced by adversarial rewriting. 

\paragraph{Length analysis.}
We first examine document length to assess whether the generated classes can be trivially
distinguished by length. Table~\ref{tab:length_stats} reports document-length statistics across datasets and generators. For XSum and arXiv, the generated texts are broadly comparable in length to their corresponding human-written sources, suggesting that length alone is unlikely to provide a dominant class
cue in these domains. ROCStory is the main exception. Because the source documents contain only five sentences, strict word-count constraints led to unnatural truncation during generation. We therefore allowed greater length flexibility for the High AI-involved and Fully AI-generated
classes. This results in larger length ratios for these classes, particularly for ROCStory. Values are computed over aligned source-generation pairs.

\begin{table*}
\centering
\resizebox{0.9\textwidth}{!}{
\begin{tabular}{llcccc}
\toprule
Dataset & Gen. & Human words & Low AI/Human & High AI/Human & Fully AI/Human \\
\midrule
XSum
& GPT
& 629.6 $\pm$ 285.0
& 0.967 $\pm$ 0.034
& 0.838 $\pm$ 0.151
& 1.012 $\pm$ 0.235 \\

XSum
& Claude
& 629.6 $\pm$ 285.0
& 0.984 $\pm$ 0.028
& 0.934 $\pm$ 0.128
& 1.206 $\pm$ 0.221 \\

arXiv
& GPT
& 758.3 $\pm$ 344.3
& 0.964 $\pm$ 0.026
& 0.884 $\pm$ 0.070
& 1.100 $\pm$ 0.244 \\

arXiv
& Claude
& 758.3 $\pm$ 344.3
& 0.944 $\pm$ 0.094
& 0.809 $\pm$ 0.098
& 0.955 $\pm$ 0.169 \\

ROCStory
& GPT
& 61.9 $\pm$ 4.2
& 1.079 $\pm$ 0.097
& 1.910 $\pm$ 0.249
& 1.659 $\pm$ 0.206 \\

ROCStory
& Claude
& 61.9 $\pm$ 4.2
& 1.180 $\pm$ 0.115
& 2.717 $\pm$ 0.388
& 1.760 $\pm$ 0.218 \\
\bottomrule
\end{tabular}}

\caption{
Length statistics across datasets and generators.
Human words reports the word count of the source texts,
while Low AI/Human, High AI/Human, and Fully AI/Human report
generated-to-source word-count ratios.
Values are mean $\pm$ standard deviation over aligned
source-generation pairs.
}
\label{tab:length_stats}
\end{table*}

\begin{figure*}
    \centering
    \includegraphics[width=1\textwidth]{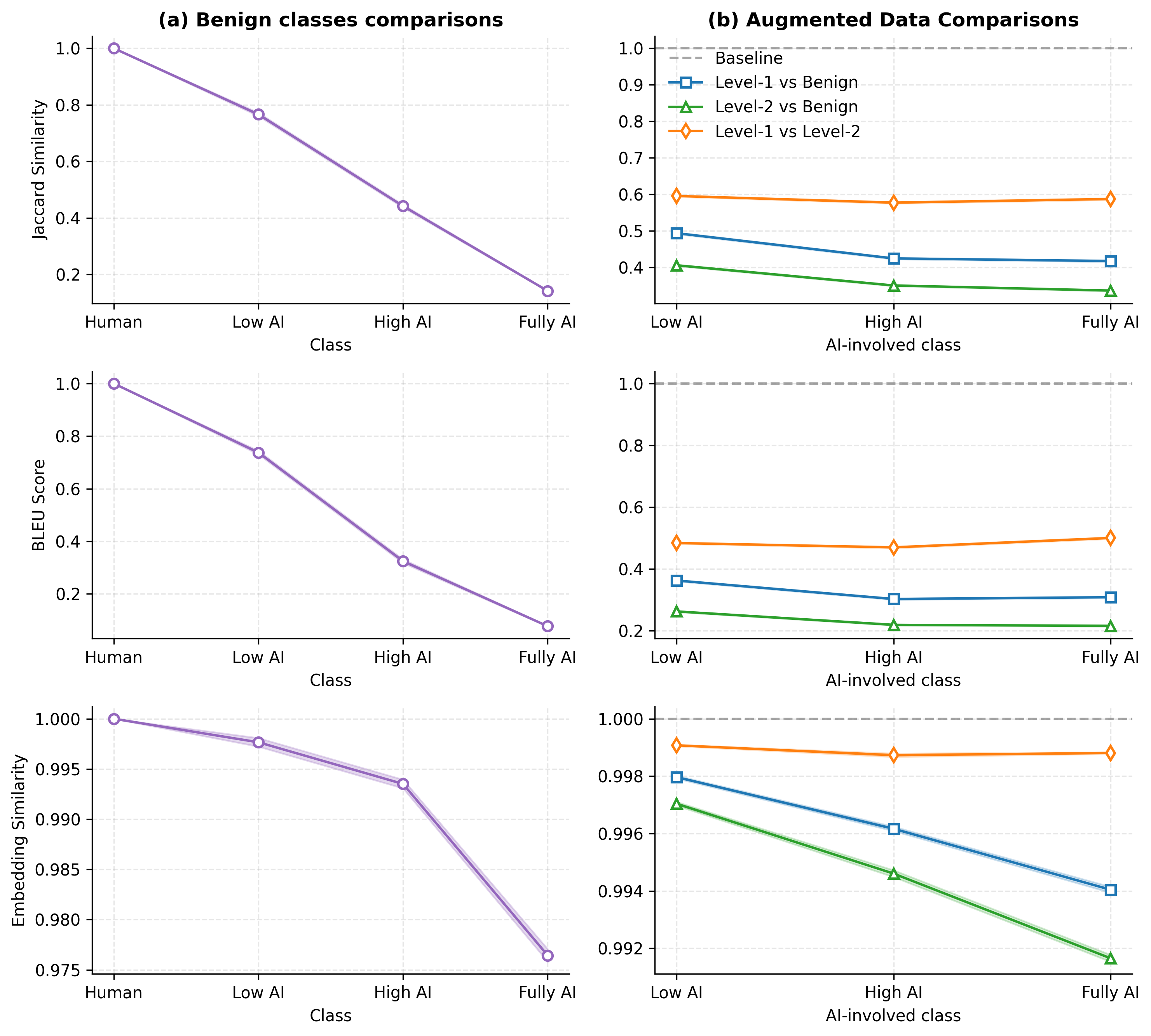}
    \caption{Quantitative similarity analysis using lexical (Jaccard), n-gram (BLEU), and semantic (embedding cosine) metrics. (a) Similarity between benign AI-involved texts and their corresponding human-written sources decreases monotonically with increasing AI involvement. (b) Adversarial paraphrases introduce substantial additional surface-level divergence from their benign counterparts while largely preserving semantic similarity.}
    \label{fig:similarity}
\end{figure*}

\paragraph{Similarity analysis.}
We next examine whether the benign AI-involvement categories exhibit the intended progression in surface form while preserving semantic content, and how adversarial paraphrasing departs from this progression. We report three complementary similarity measures. Jaccard similarity captures lexical overlap based on word-level token sets. BLEU measures n-gram overlap at the sentence level with smoothing. Embedding cosine similarity measures semantic relatedness using representations from a pretrained Longformer encoder. Figure~\ref{fig:similarity} summarizes similarity trends computed on aligned sample pairs drawn from the same underlying source documents, with mean values reported and shaded bands indicating the standard error of the mean (SEM).

Figure~\ref{fig:similarity}a compares benign AI-involved texts with their corresponding human-written sources. Across all three metrics, similarity decreases monotonically as AI involvement increases. Lexical and n-gram overlap drop substantially from Low to High AI involvement, while embedding similarity decreases more gradually. This pattern indicates increasing divergence in surface form while semantic content remains comparatively stable, consistent with the intended progression of AI editing strength.

Figure~\ref{fig:similarity}b compares adversarial paraphrases with their benign counterparts within the same AI-involvement level, as well as Adv Moderate versus Adv Strong. Relative to benign text, both adversarial settings lead to pronounced additional drops in Jaccard and BLEU, indicating substantial perturbations at the lexical and local phrasal levels. In contrast, embedding cosine similarity remains close to that of benign samples, with only minor decreases even under multi-step paraphrasing. This pattern is consistent with the design of DIPPER, which aims to evade surface-form–based detection by introducing controlled lexical substitutions and local reordering while preserving the underlying meaning. Moreover, the limited gap between Adv Moderate and Adv Strong in embedding similarity suggests diminishing semantic changes from additional paraphrasing steps. Together, these results indicate that DIPPER paraphrases introduce an extra surface-level shift that is not aligned with the natural progression of AI involvement, supporting our choice to treat DIPPER outputs as adversarial augmentations rather than redefining AI-involvement categories.


\begin{table*}[!t]
\centering
\begin{tabular}{llcc}
\toprule
Data & Setting & Original (\%) & Recalibrated (\%) \\
\midrule
XSum-GPT-4o & Benign  & 47.8 & 95.8 \\
XSum-GPT-4o & Level-1 & 51.9 & 84.1 \\
XSum-Claude  & Benign  & 48.6 & 92.8 \\
XSum-Claude  & Level-1 & 51.9 & 78.9 \\
\bottomrule
\end{tabular}
\caption{
Alignment between EditLens similarity-based categories and
our process-defined AI-involvement labels using the original
and recalibrated thresholds.
}
\label{tab:editlens_alignment}
\vspace{2em}
\begin{tabular}{llccc|ccc}
\toprule
& & \multicolumn{3}{c}{EditLens} &
\multicolumn{3}{c}{Ours} \\
\cmidrule(lr){3-5}
\cmidrule(lr){6-8}
Data & Test & Acc. & F1 & AUC & Acc. & F1 & AUC \\
\midrule
XSum-GPT-4o & Benign
& 0.950 & 0.950 & 0.991
& 0.978 & 0.978 & \textbf{0.998} \\

XSum-GPT-4o & Level-1
& 0.879 & 0.877 & 0.979
& 0.938 & 0.943 & \textbf{0.991} \\

XSum-GPT-4o & GPT-5.1
& 0.808 & 0.802 & 0.955
& 0.872 & 0.872 & \textbf{0.964} \\

XSum-Claude & Benign
& 0.861 & 0.858 & 0.984
& 0.975 & 0.975 & \textbf{0.999} \\

XSum-Claude & Level-1
& 0.778 & 0.792 & 0.940
& 0.943 & 0.944 & \textbf{0.996} \\

XSum-Claude & GPT-5.1
& 0.747 & 0.721 & 0.935
& 0.803 & 0.785 & \textbf{0.953} \\
\bottomrule
\end{tabular}
\caption{
Controlled comparison with EditLens under Level-1 training
using the same RoBERTa-large backbone.
}
\label{tab:editlens_results}
\end{table*}

\section{Additional Experiment Results}\label{additional_results}
\paragraph{Class-wise sensitivity and where robustness improves.}
Per-class F1 shows that the robustness gains are driven mainly by the ambiguous Low AI and High AI classes. On XSum (Table~\ref{tab:xsum_full_results}), benign-only training degrades sharply under paraphrased testing (e.g., on GPT-4o-generated XSum, High AI drops from 0.9888 under benign testing to 0.6567/0.4916 under Level-1/Level-2 testing), and transfer to GPT-5.1 can be especially brittle. After incorporating Level-1 adversarial data during training, Low AI and High AI become markedly more stable across testing levels and the held-out generator, indicating improved robustness to surface variation.

In contrast, the Fully AI class is consistently the easiest: its F1 remains near-saturated across conditions, suggesting that strong fully-synthetic signals persist even after paraphrasing. The Human class also benefits from adversarial exposure, but the dominant robustness gains come from reducing confusion between Low AI and High AI under distribution shift.

Similar patterns hold across domains: improvements on ROCStory concentrate in Human and Low AI, whereas arXiv again improves most for Low AI and High AI. Overall, single-step adversarial training primarily strengthens the detector on the most confusable intermediate classes, with diminishing returns from additional multi-step paraphrasing.

\paragraph{Comparison with continuous scoring.}
We further compare our formulation with EditLens~\citep{thai2026editlens}, a representative continuous estimation approach for fine-grained human-AI detection, on XSum using GPT-4o and Claude generations. We focus on XSum to avoid input-length truncation effects and use the released RoBERTa-large implementation of EditLens.

Since EditLens defines its categories using similarity-based thresholds, we recalibrate the thresholds on our training data to improve alignment with our process-defined labels (Table~\ref{tab:editlens_alignment}). We then compare the two methods under the same RoBERTa-large backbone. As shown in Table~\ref{tab:editlens_results}, our formulation achieves consistently higher performance across the evaluated benign, adversarial, and held-out rewriting settings.


\begin{table*}[!t]
  \centering
  \resizebox{0.9\textwidth}{!}{
      \begin{tabular}{c|c|cccc|ccc}
        \toprule
        \multicolumn{9}{c}{\textbf{Dataset: XSum}}\\    
        \midrule
        \multicolumn{2}{c}{\textbf{Settings}} 
        & \multicolumn{4}{c}{\textbf{F1 per Class}} 
        & \multicolumn{3}{c}{\textbf{Overall Metrics}} \\
        \textbf{Train} & \textbf{Test} & \textbf{Human} & \textbf{Low AI} & \textbf{High AI} & \textbf{Fully AI} 
        & \textbf{Acc} & \textbf{Macro F1} & \textbf{AUC} \\
        \midrule
        \multicolumn{9}{c}{\textbf{(a) Same generator and same attack.}}\\
        \midrule
        \multicolumn{9}{c}{Part I: Benign AI-involved Data generated by  GPT-4o} \\
        \midrule
        \multirow{4}{*}{Benign}  & Benign  & 0.9775 & 0.9721 & 0.9888 & 0.9836 & 0.9805 & 0.9805 & 0.9998 \\
                & Level-1 & 0.6824 & 0.7151 & 0.6567 & 0.9697 & 0.7695 & 0.7560 & 0.9386 \\
                & Level-2 & 0.4833 & 0.6131 & 0.4916 & 0.9586 & 0.6652 & 0.6367 & 0.8988 \\
                & GPT-5.1 & 0.6905 & 0.4767 & 0.0000 & 0.9836 & 0.6212 & 0.5377 & 0.8413 \\
        \midrule
        \multirow{4}{*}{Level-1} & Benign & 0.9773 & 0.9780 & 0.9945 & 0.9944 & 0.9861 & 0.9861 & 0.9990 \\
                & Level-1 & 0.9773 & 0.9143 & 0.9223 & 0.9916 & 0.9475 & 0.9514 & 0.9947~$\uparrow$ \\
                & Level-2 & 0.9773 & 0.8931 & 0.8980 & 0.9907 & 0.9321 & 0.9398 & 0.9921~$\uparrow$ \\
                & GPT-5.1 & 0.9829 & 0.7089 & 0.7864 & 0.9944 & 0.8691 & 0.8682 & 0.9662~$\uparrow$ \\
        \midrule
        \multirow{4}{*}{Level-2} & Benign & 0.9551 & 0.9379 & 0.9724 & 0.9890 & 0.9638 & 0.9636 & 0.9973 \\
                & Level-1 & 0.9551 & 0.9070 & 0.9194 & 0.9835 & 0.9396 & 0.9413 & 0.9920 \\
                & Level-2 & 0.9551 & 0.8863 & 0.8936 & 0.9817 & 0.9244 & 0.9292 & 0.9890 \\
                & GPT-5.1 & 0.9770 & 0.6892 & 0.8056 & 1.0000 & 0.8719 & 0.8680 & 0.9554 \\
        \midrule
        \multicolumn{8}{c}{Part II: Benign AI-involved Data generated by Claude Sonnet 4} \\
        \midrule
        \multirow{4}{*}{Benign}  & Benign  & 0.9663 & 0.9667 & 0.9831 & 0.9836 & 0.9749 & 0.9729 & 0.9979 \\
                & Level-1 & 0.4873 & 0.6689 & 0.6567 & 0.9053 & 0.6789 & 0.6796 & 0.8999 \\
                & Level-2 & 0.3127 & 0.5217 & 0.4916 & 0.8445 & 0.5373 & 0.5426 & 0.8326 \\
                & GPT-5.1 & 0.4886 & 0.0000 & 0.0000 & 0.9890 & 0.4903 & 0.3694 & 0.7882 \\
        \midrule
        \multirow{4}{*}{Level-1} & Benign & 0.9609 & 0.9663 & 0.9718 & 0.9783 & 0.9694 & 0.9693 & 0.9948 \\
                & Level-1 & 0.9609 & 0.9395 & 0.9282 & 0.9730 & 0.9491 & 0.9504 & 0.9927~$\uparrow$ \\
                & Level-2 & 0.9609 & 0.9151 & 0.8995 & 0.9711 & 0.9321 & 0.9367 & 0.9890~$\uparrow$ \\
                & GPT-5.1 & 0.9829 & 0.5865 & 0.7032 & 0.9424 & 0.8134 & 0.8038 & 0.9433~$\uparrow$ \\
        \midrule
        \multirow{4}{*}{Level-2} & Benign & 0.9565 & 0.9480 & 0.9659 & 0.9730 & 0.9610 & 0.9609 & 0.9960 \\
                & Level-1 & 0.9565 & 0.9235 & 0.9076 & 0.9549 & 0.9332 & 0.9356 & 0.9926 \\
                & Level-2 & 0.9514 & 0.9002 & 0.8780 & 0.9490 & 0.9143 & 0.9197 & 0.9913~$\uparrow$ \\
                & GPT-5.1 & 0.9944 & 0.5231 & 0.6761 & 0.9091 & 0.7911 & 0.7756 & 0.9324 \\
        \midrule
        \multicolumn{9}{c}{\textbf{(b) Transfer across generators}}\\
        \midrule
        \multicolumn{9}{c}{Train: GPT → Test: Claude} \\
        \midrule
        Benign  & \multirow{3}{*}{Claude} & 0.9255&0.9176&0.9551&0.9560 & 0.9387 & 0.9386 & 0.9960 \\
        Level-1 & & 0.8912&0.8673&0.8989&0.9600 & 0.9062 & 0.9043 & 0.9898 \\
        Level-2 & & 0.8718&0.8429&0.8662&0.9586 & 0.8877 & 0.8849 & 0.9833 \\
        \midrule
        \multicolumn{9}{c}{Train: Claude → Test: GPT} \\
        \midrule
        Benign  & \multirow{3}{*}{GPT-4o} & 0.9198&0.9070&0.8750&0.8945 & 0.8997&0.8991&0.9933 \\
        Level-1 & & 0.9198&0.8703&0.8864&0.9570 & 0.9078 & 0.9084 & 0.9825 \\
        Level-2 & & 0.9362&0.8750&0.8544&0.9385 & 0.8954 & 0.9010 & 0.9816 \\     
        \bottomrule
      \end{tabular}}
  \caption{
  Results on the XSum dataset. Upward arrows ($\uparrow$) indicate improved AUC relative to the immediately preceding training level under the same testing condition.
  }
  \label{tab:xsum_full_results}
\end{table*}


\label{story_full_results}
\begin{table*}[t]
  \centering
  \resizebox{0.9\textwidth}{!}{
      \begin{tabular}{c|c|cccc|ccc}
        \toprule
        \multicolumn{9}{c}{\textbf{Dataset: ROCStory}}\\    
        \midrule
        \multicolumn{2}{c}{\textbf{Settings}} 
        & \multicolumn{4}{c}{\textbf{F1 per Class}} 
        & \multicolumn{3}{c}{\textbf{Overall Metrics}} \\
        \textbf{Train} & \textbf{Test} & \textbf{Human} & \textbf{Low AI} & \textbf{High AI} & \textbf{Fully AI} 
        & \textbf{Acc} & \textbf{Macro F1} & \textbf{AUC} \\
        \midrule
        \multicolumn{9}{c}{\textbf{(a) Same generator and same attack.}}\\
        \midrule
        \multicolumn{9}{c}{Part I: Benign AI-involved Data generated by GPT-4o} \\
        \midrule
        \multirow{4}{*}{Benign}  
            & Benign  & 0.9467 & 0.9468 & 0.9834 & 0.9889 & 0.9666 & 0.9665 & 0.9996 \\
            & Level-1 & 0.8122 & 0.8604 & 0.8829 & 0.9019 & 0.8712 & 0.8644 & 0.9793 \\
            & Level-2 & 0.6452 & 0.7649 & 0.8238 & 0.8631 & 0.7953 & 0.7743 & 0.9583 \\
            & GPT-5.1 & 0.8163 & 0.5000 & 0.4310 & 0.6496 & 0.6323 & 0.5992 & 0.9062 \\
        \midrule
        \multirow{4}{*}{Level-1} 
            & Benign  & 0.8944 & 0.9026 & 0.9832 & 0.9836 & 0.9415 & 0.9410 & 0.9977 \\
            & Level-1 & 0.8889 & 0.9305 & 0.9435 & 0.9457 & 0.9332 & 0.9272 & 0.9943~$\uparrow$ \\
            & Level-2 & 0.8780 & 0.9355 & 0.9242 & 0.9307 & 0.9255 & 0.9171 & 0.9892~$\uparrow$ \\
            & GPT-5.1 & 0.8090 & 0.5139 & 0.5426 & 0.6742 & 0.6518 & 0.6349 & 0.9249~$\uparrow$ \\
        \midrule
        \multirow{4}{*}{Level-2} 
            & Benign  & 0.9157 & 0.9043 & 0.9497 & 0.9622 & 0.9331 & 0.9330 & 0.9931 \\
            & Level-1 & 0.9048 & 0.9286 & 0.9226 & 0.9284 & 0.9237 & 0.9211 & 0.9904 \\
            & Level-2 & 0.8994 & 0.9375 & 0.8992 & 0.9069 & 0.9132 & 0.9108 & 0.9876 \\
            & GPT-5.1 & 0.8172 & 0.4511 & 0.4667 & 0.6380 & 0.6212 & 0.5933 & 0.8983 \\
        \midrule
        \multicolumn{9}{c}{Part II: Benign AI-involved Data generated by Claude Sonnet 4} \\
        \midrule
        \multirow{4}{*}{Benign}  
            & Benign  & 1.0000 & 1.0000 & 0.9945 & 0.9944 & 0.9972 & 0.9972 & 1.0000 \\
            & Level-1 & 0.8357 & 0.8869 & 0.9069 & 0.8987 & 0.8871 & 0.8821 & 0.9848 \\
            & Level-2 & 0.6642 & 0.7872 & 0.8251 & 0.8275 & 0.7920 & 0.7760 & 0.9611 \\
            & GPT-5.1 & 0.8357 & 0.1443 & 0.9480 & 0.7574 & 0.7437 & 0.6714 & 0.9250 \\
        \midrule
        \multirow{4}{*}{Level-1} 
            & Benign  & 0.9770 & 0.9783 & 1.0000 & 1.0000 & 0.9889 & 0.9888 & 1.0000 \\
            & Level-1 & 0.9605 & 0.9694 & 0.9749 & 0.9587 & 0.9666 & 0.9659 & 0.9973~$\uparrow$ \\
            & Level-2 & 0.9290 & 0.9497 & 0.9605 & 0.9324 & 0.9455 & 0.9429 & 0.9951~$\uparrow$ \\
            & GPT-5.1 & 0.8854 & 0.5079 & 0.9831 & 0.8072 & 0.8189 & 0.7959 & 0.9453~$\uparrow$ \\
        \midrule
        \multirow{4}{*}{Level-2} 
            & Benign  & 0.9405 & 0.9474 & 0.9890 & 0.9888 & 0.9666 & 0.9664 & 1.0000 \\
            & Level-1 & 0.9349 & 0.9565 & 0.9699 & 0.9551 & 0.9571 & 0.9541 & 0.9966 \\
            & Level-2 & 0.9294 & 0.9510 & 0.9617 & 0.9318 & 0.9466 & 0.9435 & 0.9944 \\
            & GPT-5.1 & 0.8927 & 0.5401 & 0.9545 & 0.7719 & 0.8022 & 0.7898 & 0.9483~$\uparrow$ \\
        \midrule
        \multicolumn{9}{c}{\textbf{(b) Transfer across generators}}\\
        \midrule
        \multicolumn{9}{c}{Train: GPT → Test: Claude} \\
        \midrule
        Benign  & \multirow{3}{*}{Claude} & 0.9467 & 0.9362 & 0.9836 & 0.9888 & 0.9638 & 0.9638 & 0.9990 \\
        Level-1 & & 0.8944 & 0.8920 & 0.9091 & 0.9171 & 0.9046 & 0.9031 & 0.9902 \\
        Level-2 & & 0.9157 & 0.9049 & 0.8946 & 0.9062 & 0.9032 & 0.9054 & 0.9849 \\
        \midrule
        \multicolumn{9}{c}{Train: Claude → Test: GPT} \\
        \midrule
        Benign  & \multirow{3}{*}{GPT-4o} & 0.9368 & 0.8953 & 0.9286 & 0.9362 & 0.9248 & 0.9242 & 0.9944 \\
        Level-1 & & 0.8808 & 0.8939 & 0.8528 & 0.8871 & 0.8792 & 0.8786 & 0.9820 \\
        Level-2 & & 0.8977 & 0.8981 & 0.8421 & 0.8454 & 0.8665 & 0.8708 & 0.9785 \\     
        \bottomrule
      \end{tabular}}
  \caption{
  Results on the ROCStory dataset. Upward arrows ($\uparrow$) indicate improved AUC relative to the immediately preceding training level under the same testing condition.
  }
  \label{tab:rocstory}
\end{table*}


\begin{table*}[t]
  \centering
  \resizebox{0.9\textwidth}{!}{
      \begin{tabular}{c|c|cccc|ccc}
        \toprule
        \multicolumn{9}{c}{\textbf{Dataset: arXiv}}\\    
        \midrule
        \multicolumn{2}{c}{\textbf{Settings}} 
        & \multicolumn{4}{c}{\textbf{F1 per Class}} 
        & \multicolumn{3}{c}{\textbf{Overall Metrics}} \\
        \textbf{Train} & \textbf{Test} & \textbf{Human} & \textbf{Low AI} & \textbf{High AI} & \textbf{Fully AI} 
        & \textbf{Acc} & \textbf{Macro F1} & \textbf{AUC} \\
        \midrule
        \multicolumn{9}{c}{\textbf{(a) Same generator and same attack.}}\\
        \midrule
        \multicolumn{9}{c}{Part I: Benign AI-involved Data generated by GPT-4o} \\
        \midrule
        \multirow{4}{*}{Benign}  
            & Benign  & 0.8235 & 0.6000 & 0.6226 & 0.8547 & 0.7286 & 0.7252 & 0.9382 \\
            & Level-1 & 0.5344 & 0.3938 & 0.4405 & 0.8544 & 0.5673 & 0.5558 & 0.8343 \\
            & Level-2 & 0.3196 & 0.2958 & 0.3363 & 0.7509 & 0.4329 & 0.4257 & 0.7437 \\
            & GPT-5.1 & 0.8333 & 0.5741 & 0.6602 & 0.9709 & 0.7538 & 0.7596 & 0.9285 \\
        \midrule
        \multirow{4}{*}{Level-1} 
            & Benign  & 0.6111 & 0.3218 & 0.5645 & 0.8696 & 0.6080 & 0.5918 & 0.8870 \\
            & Level-1 & 0.6111 & 0.3467 & 0.5537 & 0.8547 & 0.6160 & 0.5916 & 0.8923~$\uparrow$ \\
            & Level-2 & 0.6111 & 0.4071 & 0.5674 & 0.8721 & 0.6393 & 0.6144 & 0.8964~$\uparrow$ \\
            & GPT-5.1 & 0.6197 & 0.2791 & 0.6364 & 0.9174 & 0.6332 & 0.6132 & 0.8761 \\
        \midrule
        \multirow{4}{*}{Level-2} 
            & Benign  & 0.7595 & 0.4884 & 0.6720 & 0.9259 & 0.7186 & 0.7115 & 0.9230~$\uparrow$ \\
            & Level-1 & 0.7595 & 0.4868 & 0.6218 & 0.8734 & 0.6905 & 0.6854 & 0.9148~$\uparrow$ \\
            & Level-2 & 0.7595 & 0.5022 & 0.6041 & 0.8451 & 0.6794 & 0.6777 & 0.9133~$\uparrow$ \\
            & GPT-5.1 & 0.7595 & 0.4000 & 0.6667 & 0.9524 & 0.7035 & 0.6947 & 0.9050~$\uparrow$ \\

        \midrule
        \multicolumn{9}{c}{Part II: Benign AI-involved Data generated by Claude Sonnet 4} \\
        \midrule
        \multirow{4}{*}{Benign}  
            & Benign  & 0.9091 & 0.8842 & 0.9505 & 0.9709 & 0.9296 & 0.9287 & 0.9913 \\
            & Level-1 & 0.4265 & 0.5316 & 0.6358 & 0.8090 & 0.5931 & 0.6007 & 0.8274 \\
            & Level-2 & 0.2521 & 0.4038 & 0.4776 & 0.6552 & 0.4228 & 0.4472 & 0.7279 \\
            & GPT-5.1 & 0.6429 & 0.1231 & 0.0392 & 0.7042 & 0.5025 & 0.3774 & 0.8934 \\
        \midrule
        \multirow{4}{*}{Level-1} 
            & Benign  & 0.8172 & 0.7312 & 0.8829 & 0.9901 & 0.8593 & 0.8554 & 0.9653 \\
            & Level-1 & 0.8172 & 0.7778 & 0.8341 & 0.9346 & 0.8481 & 0.8409 & 0.9625~$\uparrow$ \\
            & Level-2 & 0.8172 & 0.7927 & 0.7948 & 0.9102 & 0.8337 & 0.8287 & 0.9571~$\uparrow$ \\
            & GPT-5.1 & 0.7917 & 0.4691 & 0.6909 & 0.9009 & 0.7286 & 0.7132 & 0.9139~$\uparrow$ \\
        \midrule
        \multirow{4}{*}{Level-2} 
            & Benign  & 0.8660 & 0.7778 & 0.8727 & 0.9703 & 0.8744 & 0.8717 & 0.9771~$\uparrow$ \\
            & Level-1 & 0.8660 & 0.8023 & 0.8372 & 0.9474 & 0.8653 & 0.8632 & 0.9698~$\uparrow$ \\
            & Level-2 & 0.8660 & 0.8090 & 0.8050 & 0.9177 & 0.8477 & 0.8494 & 0.9646~$\uparrow$ \\
            & GPT-5.1 & 0.7925 & 0.4500 & 0.6796 & 0.8991 & 0.7236 & 0.7053 & 0.9224~$\uparrow$ \\
                \midrule
        \multicolumn{9}{c}{\textbf{(b) Transfer across generators}}\\
        \midrule
        \multicolumn{9}{c}{Train: GPT → Test: Claude} \\
        \midrule
        Benign  & \multirow{3}{*}{Claude} & 0.7692 & 0.5349 & 0.6847 & 0.8727 & 0.7236 & 0.7154 & 0.9361 \\
        Level-1 & & 0.5946 & 0.4403 & 0.6276 & 0.8761 & 0.6619 & 0.6346 & 0.9042 \\
        Level-2 & & 0.7059 & 0.5067 & 0.6590 & 0.8772 & 0.7034 & 0.6872 & 0.9245 \\
        \midrule
        \multicolumn{9}{c}{Train: Claude → Test: GPT} \\
        \midrule
        Benign  & \multirow{3}{*}{GPT-4o} & 0.8738 & 0.6218 & 0.2456 & 0.7731 & 0.6784 & 0.6286 & 0.9230 \\
        Level-1 & & 0.8172 & 0.7128 & 0.6912 & 0.8500 & 0.7593 & 0.7678 & 0.9251 \\
        Level-2 & & 0.8485 & 0.6714 & 0.6287 & 0.8211 & 0.7174 & 0.7424 & 0.9084 \\ 
        \bottomrule
      \end{tabular}}
      \caption{
      Results on the Arxiv dataset. Upward arrows ($\uparrow$) indicate improved AUC relative to the immediately preceding training level under the same testing condition.
      }
      \label{tab:arxiv}
\end{table*}


\end{document}